\pdfoutput=1

\documentclass[11pt]{article}

\usepackage[final]{acl}

\usepackage{times}
\usepackage{latexsym}

\usepackage[T1]{fontenc}

\usepackage[utf8]{inputenc}

\usepackage{microtype}

\usepackage{inconsolata}

\usepackage{graphicx}

\usepackage{amsmath}
\usepackage{amsfonts}
\usepackage{xspace}
\usepackage{xparse}
\usepackage{booktabs}
\usepackage{multirow}
\usepackage{array}
\usepackage{arydshln}
\usepackage[dvipsnames]{xcolor}

\usepackage{tikz}
\usetikzlibrary{shadows, fit, shapes, arrows.meta, positioning, calc}
\usepackage{setspace}
\usepackage{enumitem}
\usepackage{arydshln}
\usepackage{subcaption}
\usepackage{soul}
\usepackage{fontawesome}
\usepackage[multiple]{footmisc}
\usepackage{tabularx}
\usepackage{adjustbox}

\newcommand\anonym[1]{}

\newcommand{\ie}{\textit{i.e.}\xspace}

\newcommand{\Wqkvo}{W_{QKVO}}

\newcommand{\orangesum}{OrangeSum\xspace}

\newcommand{\wikilargefr}{WikiLarge FR\xspace}

\newcommand{\falcdataset}{ETR-fr\xspace}
\newcommand{\falcdatasetpolitic}{\falcdataset-politic\xspace}
\newcommand{\falc}{ETR\xspace}

\newcommand{\mistral}{Mistral-7B\xspace}

\newcommand{\llamatr}{LlaMA-3-8B\xspace}

\newcommand{\llamaVIII}{\llamatr}
\newcommand{\llamaVIIIinst}{\llamatr}
\newcommand{\mistVII}{\mistral}
\newcommand{\mistVIIinst}{\mistral}
\newcommand{\deepseekrI}{DeepSeek-R1-8B\xspace}
\newcommand{\rag}{RAG}
\newcommand{\zeroshot}{Zero-Shot}
\newcommand{\COT}{CoT}

\newcommand{\novelty}{Novelty}

\newcommand{\lora}{LoRA\xspace}
\newcommand{\mtllora}{MTL-LoRA\xspace}

\newcommand{\fscore}{F1-score\xspace}
\newcommand{\bleu}{BLEU\xspace}
\newcommand{\rouge}{ROUGE\xspace}
\newcommand{\rone}{\rouge-1\xspace}
\newcommand{\RI}{R-1 $\uparrow$\xspace}
\newcommand{\rtwo}{\rouge-2\xspace}
\newcommand{\RII}{R-2 $\uparrow$\xspace}
\newcommand{\rl}{\rouge-L\xspace}
\newcommand{\RL}{R-L $\uparrow$\xspace}

\newcommand{\bertscore}{BERTScore\xspace}
\newcommand{\bertscoref}{\bertscore-F1\xspace}

\newcommand{\bertf}{BERT-F1 $\uparrow$\xspace}
\newcommand{\sari}{SARI $\uparrow$\xspace}

\newcommand{\kmre}{KMRE\xspace}

\newcommand{\SRB}{SRB $\uparrow$\xspace}
\newcommand{\compratio}{Comp. ratio\xspace}

\definecolor{box}{HTML}{f5a835}

\newcommand{\wordchoice}{Words}
\newcommand{\infochoice}{Informations}
\newcommand{\sentence}{Sentences}
\newcommand{\illustrations}{Illustrations}
\newcommand{\globalscore}{Global}

\newcommand{\fluency}{Fluency}
\newcommand{\grammar}{Grammar}
\newcommand{\relevance}{Relevance}
\newcommand{\coherence}{Coherence}
\newcommand{\quality}{Overall Quality}

\newcommand{\nsamples}{\# Examples\xspace}
\newcommand{\nsents}{\# Sentences\xspace}
\newcommand{\nwords}{\# Words\xspace}
\newcommand{\sentslen}{Sentence length\xspace}
\newcommand{\source}{source\xspace}
\newcommand{\target}{target\xspace}

\newcolumntype{L}{>{\bfseries}l}%

\title{
    Facilitating Cognitive Accessibility with LLMs:\\A Multi-Task Approach to Easy-to-Read Text Generation
}
\author{
 \textbf{François Ledoyen\textsuperscript{1,2}},
 \textbf{Gaël Dias\textsuperscript{1}},
 \textbf{Jérémie Pantin\textsuperscript{1}},
 \textbf{Alexis Lechervy\textsuperscript{1}},
 \\
 \textbf{Fabrice Maurel\textsuperscript{1}},
 \textbf{Youssef Chahir\textsuperscript{1}}
\\
 \textsuperscript{1}Université Caen Normandie, ENSICAEN, CNRS, Normandie Univ,
 \\GREYC UMR 6072, F-14000 Caen, France \\
 \textsuperscript{2}Koena SAS, F-31450 Fourquevaux, France
\\
 \small{
   \textbf{Correspondence:} \href{mailto:ledoyenfrancois@gmail.com}{ledoyenfrancois@gmail.com}
 }
}

\begin{document}
\maketitle

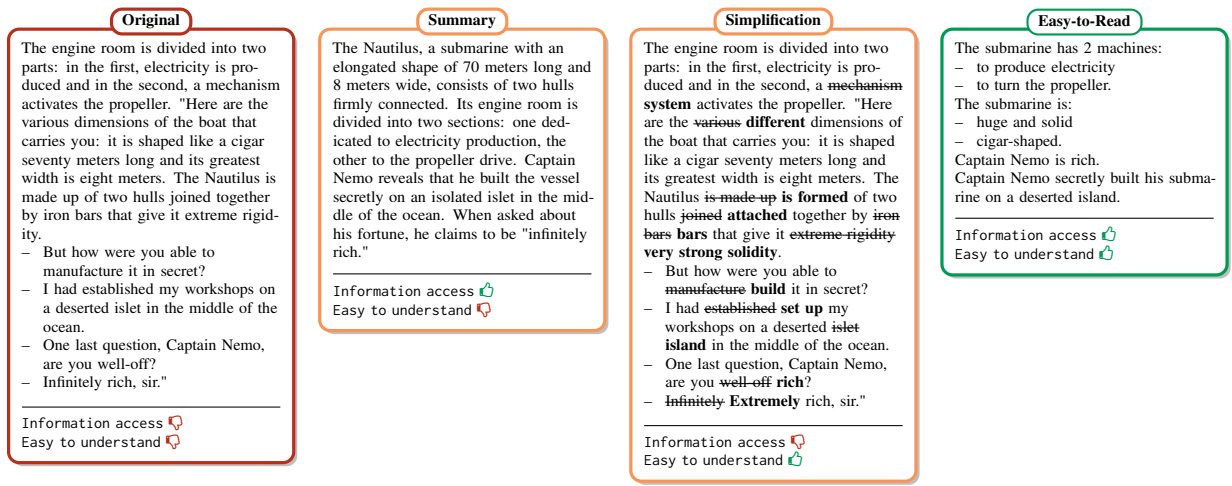
\begin{figure*}[h!]
\centering
\tiny
\begin{tikzpicture}[
    node distance=0.3cm and 0.3cm,
    box/.style={
        draw, 
        rounded corners, 
        fill=white, 
        text width=3.4cm, 
        very thick,
        shadow scale=1,
        drop shadow, 
        inner sep=5pt   
    },
    title/.style={
        draw, 
        very thick,
        rounded corners, 
        fill=white, 
        font=\bfseries,
        yshift=-0.5em,
        inner xsep=4pt,
    }
]
\newcommand{\tup}{{\textcolor{ForestGreen}{\faThumbsOUp}}}
\newcommand{\tdown}{{\textcolor{BrickRed}{\faThumbsODown}}}
\newcommand{\orgcolor}{BrickRed}
\newcommand{\summcolor}{Peach}
\newcommand{\simplcolor}{Peach}
\newcommand{\plainlanguage}{Peach}
\newcommand{\etrcolor}{ForestGreen}

\newcommand{\infoblock}[2]{
  \vspace{1em}
  \hrule
  \vspace{0.5em}
  \texttt{Information access} #1 \\
  \texttt{Easy to understand} #2
}
\node[box, draw=\orgcolor] (original) at (0,0) {
The engine room is divided into two parts: in the first, electricity is produced and in the second, a mechanism activates the propeller.
"Here are the various dimensions of the boat that carries you: it is shaped like a cigar seventy meters long and its greatest width is eight meters.
The Nautilus is made up of two hulls joined together by iron bars that give it extreme rigidity.
\begin{itemize}[noitemsep, topsep=0pt, leftmargin=*, label={--}]
\item But how were you able to manufacture it in secret?
\item I had established my workshops on a deserted islet in the middle of the ocean.
\item One last question, Captain Nemo, are you well-off?
\item Infinitely rich, sir."
\end{itemize}
\infoblock{\tdown}{\tdown}
};
\node[title, above=0cm of original, draw=\orgcolor] (orgtitle) {Original};

\node[box, right=of original.north east, anchor=north west, draw=\summcolor] (summary) {
The Nautilus, a submarine with an elongated shape of 70 meters long and 8 meters wide, consists of two hulls firmly connected. Its engine room is divided into two sections: one dedicated to electricity production, the other to the propeller drive. Captain Nemo reveals that he built the vessel secretly on an isolated islet in the middle of the ocean. When asked about his fortune, he claims to be "infinitely rich."
\infoblock{\tup}{\tdown}
};
\node[title, above=0cm of summary, draw=\summcolor] (summtitle) {Summary};

\node[box, right=of summary.north east, anchor=north west, draw=\simplcolor] (simplification) {
The engine room is divided into two parts: in the first, electricity is produced and in the second, a \textst{mechanism} \textbf{system} activates the propeller.
"Here are the \textst{various} \textbf{different} dimensions of the boat that carries you: it is shaped like a cigar seventy meters long and its greatest width is eight meters.
The Nautilus \textst{is made up} \textbf{is formed} of two hulls \textst{joined} \textbf{attached} together by \textst{iron bars} \textbf{bars} that give it \textst{extreme rigidity} \textbf{very strong solidity}.
\begin{itemize}[noitemsep, topsep=0pt, leftmargin=*, label={--}]
    \item But how were you able to \textst{manufacture} \textbf{build} it in secret?
    \item I had \textst{established} \textbf{set up} my workshops on a deserted \textst{islet} \textbf{island} in the middle of the ocean.
    \item One last question, Captain Nemo, are you \textst{well-off} \textbf{rich}?
    \item \textst{Infinitely} \textbf{Extremely} rich, sir."
\end{itemize}
\infoblock{\tdown}{\tup}
};
\node[title, above=0cm of simplification, draw=\simplcolor] (simpletitle) {Simplification};



\node[box, right=of simplification.north east, anchor=north west, draw=\etrcolor] (etr) {
The submarine has 2 machines:
\begin{itemize}[noitemsep, topsep=0pt, leftmargin=*, label={--}]
    \item to produce electricity
    \item to turn the propeller.
\end{itemize}
The submarine is:
\begin{itemize}[noitemsep, topsep=0pt, leftmargin=*, label={--}]
\item huge and solid
\item cigar-shaped.
\end{itemize}
Captain Nemo is rich.\\
Captain Nemo secretly built his submarine
on a deserted island.
\infoblock{\tup}{\tup}
};
\node[title, above=0cm of etr, thick, draw=\etrcolor] (etrtitle) {Easy-to-Read};


\end{tikzpicture}
\caption{
    Different versions derived from a passage of \textit{Twenty Thousand Leagues Under the Sea} by Jules Verne: 
    from left to right, 
    the original passage, 
    an abstractive summary, 
    a lexical simplification (crossed-out followed by words in bold indicate substitutions), 
    and
    an Easy-to-Read transcription targeting readers with cognitive impairment. 
}
\label{fig:illustation}
\end{figure*}

\begin{abstract}
Simplifying complex texts is essential to ensure equitable access to information, particularly for individuals with cognitive impairments.  
The Easy-to-Read (ETR) initiative provides a framework to make content more accessible for these individuals.  
However, manually creating such texts remains time-consuming and resource-intensive.  
In this work, we investigate the potential of large language models (LLMs) to automate the generation of ETR content.  
To address the scarcity of aligned corpora and the specific constraints of ETR, we propose a multi-task learning (MTL) approach that trains models jointly on text summarization, text simplification, and ETR generation.  
We explore two complementary strategies: multi-task retrieval-augmented generation (RAG) for in-context learning (ICL), and MTL-LoRA for parameter-efficient fine-tuning (PEFT).  
Our experiments with Mistral-7B and LLaMA-3-8B, conducted on ETR-fr, a new high-quality dataset, show that MTL-LoRA consistently outperforms all other strategies in in-domain settings, while the MTL-RAG-based approach achieves better generalization in out-of-domain scenarios.
Our code is publicly available at \url{https://github.com/FrLdy/ETR-PEFT-Composition}.  
\end{abstract}

\section{Introduction}

Mental health conditions and intellectual disabilities affect millions of people worldwide, creating significant challenges for equitable access to information~\cite{MAULIK2011419, GUSTAVSSON2011718}.  
These individuals often struggle with complex texts, which limits their participation in healthcare, education, and civic life.  
Despite international initiatives for inclusion\footnote{\href{https://sdgs.un.org/goals}{UN Sustainable Development Goals}}\textsuperscript{,}\footnote{\href{https://unsdg.un.org/2030-agenda/universal-values/leave-no-one-behind}{Leave No One Behind Principle}},  
accessible written content remains a major barrier to full participation.  

While Easy-to-Read (ETR) \cite{pathways_information_2021}, text simplification \cite{paetzold_unsupervised_2016}, summarization \cite{rush_neural_2015}, and plain language \cite{maass2020easy} all aim to improve comprehension, they differ in purpose, audience, and methods.
Text simplification rewrites content to enhance readability while preserving the original informational content \cite{gooding-2022-ethical, stajner-2021-automatic}.
Summarization shortens the original text by extracting and presenting only the key points, often without rewording for greater clarity \cite{rush_neural_2015}.
Plain language addresses broad audiences, including people with limited literacy, by using clear structure and simple vocabulary, but it may still be too complex for individuals with cognitive impairments \cite{maass2020easy}.
ETR, by contrast, is a rigorously standardized form of text adaptation developed specifically for individuals with intellectual disabilities. It requires strict adherence to \citet{pathways_information_2021} guidelines, which mandate very short sentences, highly simplified vocabulary, visual supports, and obligatory end-user testing. The primary goal is to foster the autonomy of readers with cognitive impairments. Importantly, ETR materials must be co-created by subject-matter experts together with individuals with cognitive disabilities to ensure compliance with ETR standards and eligibility for European ETR certification\footnote{\url{https://www.inclusion-europe.eu/wp-content/uploads/2021/02/How-to-use-ETR-logo..pdf}}.

However, ETR adoption remains limited due to the time-consuming and costly nature of manual adaptation, coupled with the lack of robust automated tools tailored to the linguistic and cognitive requirements of ETR content~\cite{chebab_falc_2019}. The potential of LLMs for improving accessibility~\cite{Freyer2024} is limited by the scarcity of high-quality, document-aligned ETR datasets.
Existing resources, such as ClearSim~\cite{espinosa-zaragoza_automatic_2023}, are limited and only partially aligned, highlighting the broader challenge of constructing or recovering document-aligned corpora suitable for model training. Consequently, prior studies \cite{martinez_exploring_2024, sun-etal-2023-exploiting} have approached the ETR task by leveraging sentence simplification or summarization resources, which fall short of fully meeting ETR-specific requirements.

In this paper, we address these gaps by introducing \falcdataset, the first dataset of 523 paragraph-aligned text pairs fully compliant with the European ETR guidelines \citep{pathways_information_2021}. 
We explore multi-task learning (MTL) to boost ETR generation by combining summarization and simplification, traditionally applied in isolation.
In particular, we evaluate two MTL strategies: in-context learning (ICL) via a multi-task variant of retrieval-augmented generation (RAG), and parameter-efficient fine-tuning (PEFT) using MTL-LoRA~\cite{mtllora}. Experiments are conducted on Mistral-7B~\cite{jiang_mistral_2023} and LLaMA-3-8B~\cite{grattafiori2024llama3herdmodels}, and compared against single-task baselines. The evaluation framework combines standard automatic metrics with detailed human assessment based on a 38-point rubric from the European ETR guidelines, measuring clarity, coherence, and accessibility. 
The experiments conducted on ETR-fr indicate that, in the majority of cases, MTL setups provide clear advantages over single-task baselines.
Furthermore, the results indicate that the MTL-RAG-based strategy supports better generalization in out-of-domain scenarios, while MTL-LoRA consistently achieves superior performance in in-domain settings. 

Our contributions are: (1) we release ETR-fr, the first high-quality, paragraph-aligned dataset for ETR generation, fully compliant with European guidelines and in the French language; (2) we benchmark MTL-RAG and MTL-LoRA approaches for ETR generation; (3) we propose a comprehensive evaluation combining automatic and human assessment based on official European ETR standards; (4) we evaluate model generalization to new domains, including political texts aimed at fostering civic engagement among individuals with cognitive disabilities.

\section{Related Work}

\paragraph{Inclusive Text Generation.}
Recent research has aimed to support communication for users with cognitive impairments, often through dialogue systems~\cite{martin_bridging_2024,murillo-morales_automatic_2020,huq_dialogue_2024,wang_simulated_2024}. Much of this work has focused on dyslexia. For example, \citet{goodman_lampost_2022} introduced an email writing assistant built on LaMDA~\cite{thoppilan_lamda_2022}, but observed that its outputs often lacked precision. In the French context, HECTOR~\cite{todirascu_hector_2022} investigated lexical and syntactic simplification, with mixed outcomes.

Similar challenges are observed across other languages. In German, several studies explore simplification for individuals with learning difficulties, though often without referencing the ETR framework~\cite{Hansen_translation_etr_2020, anschutz-etal-2023-language, deilen-etal-2023-using, stodden-etal-2023-deplain}. For English, relevant work includes \citet{yaneva-2015-easy}. 
In Finnish, \citet{dmitrieva_towards_2024} trained mBART~\cite{liu_multilingual_2020} and FinGPT~\cite{luukkonen_fingpt_2023} on the automatically aligned Easy-Finnish dataset, though text pairings may be inaccurate and the data does not fully follow ETR guidelines.
In Spanish, ClearText~\cite{espinosa-zaragoza_automatic_2023} leverages ChatGPT to simplify administrative texts, though its corpus remains limited and prone to errors. Additionally, \citet{martinez_exploring_2024} constructed a sentence-level simplification dataset and fine-tuned LLaMA-2~\cite{touvron_llama_2023}, revealing that translation-based methods are vulnerable to semantic drift and domain mismatches.

\begin{table*}[ht!]
\tiny
\centering

\renewcommand{\arraystretch}{1.2}
\resizebox{\textwidth}{!}{
\begin{tabular}{Lcccccccccccc}
\toprule
 & \multirow[c]{2}[1]{*}{\textbf{\nsamples}} & \multicolumn{2}{c}{\textbf{\nwords}} & \multicolumn{2}{c}{\textbf{\nsents}} & \multicolumn{2}{c}{\textbf{\sentslen}} & \multicolumn{2}{c}{\textbf{\kmre $\uparrow$}} & \multirow[c]{2}[1]{*}{\textbf{\novelty\ (\%)}} & \multirow[c]{2}[1]{*}{\textbf{\compratio (\%)}} \\
 \cmidrule(lr){3-4} \cmidrule(lr){5-6} \cmidrule(lr){7-8} \cmidrule(lr){9-10}
 & \textbf{} & \textbf{\source} & \textbf{\target} & \textbf{\source} & \textbf{\target} & \textbf{\source} & \textbf{\target} & \textbf{\source} & \textbf{\target} & \textbf{} & \textbf{} \\
\midrule
\textbf{\falcdataset} & $523$ & $102.76$ & $46.15$ & $9.30$ & $7.13$ & $12.57$ & $7.89$ & $91.43$ & $98.94$ & $53.80$ & $50.05$ \\
\textbf{Train} & $399$ & $99.70$ & $46.50$ & $8.92$ & $7.48$ & $12.57$ & $6.92$ & $91.03$ & $99.71$ & $53.79$ & $49.04$ \\
\textbf{Dev} & $71$ & $100.76$ & $48.59$ & $9.03$ & $7.77$ & $13.59$ & $6.90$ & $89.50$ & $100.59$ & $52.96$ & $44.47$ \\
\textbf{Test} & $53$ & $128.47$ & $40.26$ & $12.51$ & $10.34$ & $11.16$ & $3.97$ & $97.02$ & $103.67$ & $55.01$ & $65.19$ \\
\midrule
\textbf{\falcdatasetpolitic} & $33$ & $96.27$ & $62.85$ & $6.03$ & $6.42$ & $16.69$ & $11.84$ & $74.00$ & $87.74$ & $63.78$ & $29.17$ \\
\midrule
\textbf{\wikilargefr} & $296{,}402$ & $34.88$ & $29.28$ & $1.68$ & $1.56$ & $27.53$ & $23.74$ & $65.38$ & $71.35$ & $31.97$ & $12.79$ \\
\textbf{\orangesum} & $24{,}401$ & $375.98$ & $34.00$ & $17.15$ & $1.86$ & $22.77$ & $21.68$ & $69.80$ & $68.32$ & $38.24$ & $89.16$ \\
\bottomrule
\end{tabular}}
\caption{
\textbf{Statistics across \falcdataset, \falcdatasetpolitic, and ETR-related tasks}, i.e. sentence simplification and text summarization with \wikilargefr and \orangesum.
Results are reported on average per document.
}
\label{tab:dataset_comparaison}
\end{table*}

\paragraph{In-Context Learning (ICL).}
ICL allows LLMs to learn tasks from examples without parameter updates~\cite{gpt3,chowdhery2022palm,openai:2023gpt4,touvron2023llamaopenefficientfoundation}. Instruction tuning and Chain-of-Thought (CoT) prompting have been shown to improve task performance and reasoning~\cite{liu2023prompt,wei2022chain,yin-etal-2023-read}.
\citet{tang-etal-2023-context} assessed ICL for controlled summarization, focusing on entity inclusion and length constraints. They observed that smaller models offered stronger controllability, while larger models achieved higher ROUGE scores. However, precise length control remained challenging.
Prompt quality and exemplar selection critically affect ICL outcomes~\cite{lu-etal-2022-fantastically,dong-etal-2024-survey}. Retrieval-augmented methods~\cite{kategpt,ram-etal-2023-context} have been proposed to improve exemplar selection. For simplification, \citet{vadlamannati-sahin-2023-metric} have used metric-based selection (e.g., SARI, BERTScore) to improve output quality.
Multi-task ICL and cross-task prompting~\cite{bhasin-etal-2024-multi,wei2023larger,chatterjee-etal-2024-language} further enhance generalization and stability, especially on unseen tasks, by leveraging format-aware prompts and semantically related exemplars.

\paragraph{PEFT for Multi-Task Learning.}
Parameter-efficient fine-tuning (PEFT) methods such as LoRA~\cite{hu2022lora}, QLoRA~\cite{dettmers2023qlora} and DoRA~\cite{liu2024dora} enable scalable adaptation of LLMs by modifying only a subset of parameters. LoRA leverages the intrinsic dimensionality of language models to achieve strong performance with minimal computational overhead.
However, LoRA-based strategies struggle in multi-task settings due to conflicting updates across tasks~\cite{wang2023multilora}. Alternatives such as MultiLoRA~\cite{wang2023multilora} and MoELoRA~\cite{Liu2023WhenMM} aim to balance generalization with task specificity, but still face challenges related to task routing and interference mitigation. To overcome these limitations, \citet{mtllora} introduced MTL-LoRA, which combines shared and task-specific modules, achieving competitive results on GLUE~\cite{wang-etal-2018-glue} with fewer trainable parameters.

\section{\falcdataset Dataset}
\label{sec:dataset}

While several datasets exist for text simplification and summarization~\cite{gala_alector_2020,hauser-etal-2022-multilingual,kamal_eddine_barthez_2021,LiuSPGSKS18}, there remains a notable lack of high-quality, document-aligned corpora for ETR generation.
To address this gap, we introduce the {\falcdataset} dataset, constructed from the François Baudez Publishing collection\footnote{\url{http://www.yvelinedition.fr/Facile-a-lire}}, which provides literature specifically designed for readers with cognitive impairments, following European ETR guidelines.
A dataset sheet~\cite{gebru_datasetsheet_2021}, outlining the data collection methodology, preprocessing steps, and distribution details, is provided in Appendix~\ref{apdx: etr-fr dataset sheet}.

\paragraph{Description.}
\falcdataset consists of 523 paragraph-aligned text pairs in French. Table~\ref{tab:dataset_comparaison} outlines key dataset statistics, including KMRE readability score~\cite{kandel_application_1958}, compression ratios \cite{kamal_eddine_barthez_2021}, and lexical novelty \citet{narayan_dont_2018}. On average, the dataset yields a compression rate of 50.05\%, with a reduction of 56.61 words and 2.17 sentences per pair. The average novelty rate is 53.80\%, reflecting the proportion of newly introduced unigrams in target texts. Readability improves by 7.51 KMRE points from source to target. 

\paragraph{Dataset Splits.}
The dataset is partitioned into fixed train, validation, and test subsets. 
The test set comprises two books selected to maximize diversity in text length, word count, sentence structure, compression, novelty, and readability. The remaining nine books are divided into training and validation sets via a stratified split. This setup was used to test hard configurations for ETR generation and ensure non-thematic and lexical overlap. 

\paragraph{ETR-fr-politic.}
To assess generalization and robustness, we introduce \falcdatasetpolitic, an out-of-domain test set with 33 ETR paragraphs sampled from the 2022 French presidential election programs, which adhere to ETR guidelines\footnote{\url{https://www.cnccep.fr/candidats.html}} and manually aligned.
Compared to \falcdataset test set, the \falcdatasetpolitic dataset features shorter source texts (96.27 vs. 128.47 words) and fewer sentences (6.03 vs. 12.51), but yields longer rewritten outputs (62.85 vs.\ 40.26 words). Additionally, \falcdatasetpolitic exhibits higher novelty (63.78\% vs.\ 55.01\%) and significantly lower compression ratios (29.17\% vs.\ 65.19\%), indicating a greater degree of content expansion. 
While \falcdataset test set exhibits higher overall simplicity scores both before and after rewriting (97.02 and 103.67) compared to \falcdatasetpolitic (74.00 and 87.74), the latter achieves a greater simplification gain, with a larger increase in \kmre (+13.74 vs.\ +6.65 points).
Overall, \falcdatasetpolitic poses a more challenging and higher-novelty setting for evaluating ETR systems in politically sensitive, real-world rewriting contexts.

\paragraph{\falcdataset vs. Related Tasks.}
Table~\ref{tab:dataset_comparaison} compares \falcdataset with two gold-standard datasets on related tasks, respectively text simplification and summarization: \wikilargefr~\cite{cardon_french_2020} and \orangesum~\cite{kamal_eddine_barthez_2021}. While \wikilargefr is larger (296K sentence pairs), it is limited to sentence-level simplification, with short inputs (34.88 words, 1.68 sentences on average). 
In contrast, \falcdataset and \orangesum support transformations at the paragraph and document levels, respectively, providing significantly longer inputs of 102.76 and 375.98 words.
\falcdataset demonstrates a balanced compression ratio (50.05\%) higher than \wikilargefr (12.79\%) but lower than the extreme summarization found in \orangesum (89.16\%). Notably, \falcdataset offers the highest lexical richness and abstraction, evidenced by its top KMRE scores (91.43 source, 98.94 target) and novelty rate (53.80\%). Simplified outputs also exhibit syntactic simplification, with shorter sentence lengths (7.89 words per sentence). 
In summary, while \wikilargefr is suited for sentence-level simplification and \orangesum for summarization, \falcdataset supports paragraph-level simplification, emphasizing lexical and structural transformation, making it well-suited for users with cognitive disabilities.

\section{Multi-Task ETR Generation}


\subsection{Datasets, LLMs and Metrics}\label{sec: dataset llms metrics}
\paragraph{Datasets.}
Our experiments leverage the \falcdataset dataset as the primary resource, supplemented by related rewriting tasks sourced from the \orangesum summarization dataset and the sentence simplification dataset \wikilargefr. 

\paragraph{Models.}
To evaluate the effectiveness of MTL for ETR transcription, we selected two recent LLMs that demonstrate strong generalization capabilities across a variety of NLP tasks : 
Llama3-8B~\cite{grattafiori2024llama3herdmodels} and Mistral-7B~\cite{jiang_mistral_2023}\footnote{We evaluated the \deepseekrI model. Its performance was notably lower than that of the other models. Results are reported in Table~\ref{table: avg results} from Appendix~\ref{apdx: qtt results}}. 
Note that foundation models are used for PEFT and their Instruct versions for ICL. 

\paragraph{Metrics.}
Since no dedicated evaluation metrics exist for ETR generation, we propose assessing it using standard summarization and text simplification metrics. 
For summarization, we report F1-scores for ROUGE-1, ROUGE-2, and ROUGE-L \cite{lin_rouge_2004}, along with BERTScore \cite{zhang_bertscore_2020}. 
For simplification, we include SARI \cite{xu_optimizing_2016}, the novelty ratio for new unigrams \cite{kamal_eddine_barthez_2021}. 
BLEU~\cite{papineni_bleu_2002} and KMRE, are excluded, as it is unsuitable for text simplification \cite{sulem_simple_2018,xu_optimizing_2016,tanprasert-kauchak-2021-flesch}.
To unify quality assessment of \falc texts, we propose SRB, a composite score combining SARI, ROUGE-L, and BERTScore-F1 via harmonic mean. This metric captures simplification, summarization, and meaning preservation for holistic ETR evaluation. 

More details about metrics and models are available in Appendix~\ref{apdx: implementation details}.

\subsection{Multi-Task In-Context Learning}

\paragraph{Single-Task Baselines.} As baseline, we evaluate three single-task in-context learning strategies: zero-shot prompting~\cite{kojima-zs-2022}, chain-of-thought prompting~\cite{wei2022chain}, and retrieval-augmented generation~\cite{lewis-rag-2020}. In the zero-shot setting, the model is provided only with ETR task-specific instructions, without any examples, serving as a baseline to assess the model’s ability to generalize purely from the prompt. To enhance reasoning in more complex tasks, we incorporate CoT prompting, which explicitly elicits intermediate reasoning steps in the prompt. 
For a fair and reproducible evaluation, we use consistent instruction-based prompt templates across all models, as detailed in Appendix~\ref{section:prompts}. 

\paragraph{Multi-Task RAG.}\label{sec:mtl_rag}
To enable few-shot multi-task ICL, we implement a multi-task RAG. Demonstrations from multiple tasks are retrieved and incorporated into the prompt.
We explore three sequencing strategies for organizing demonstrations within the prompt context, which are listed as follows.

\begin{description}
[leftmargin=0pt, font=\normalfont\itshape]

\item[Random Ordering:] Examples from all 3 tasks are interleaved in a fully randomized manner (e.g., $t_1, t_3, t_3, t_2, t_1, t_1, t_3, t_2, t_2$), serving as a baseline to assess robustness to prompt structure.

\item[Task-Grouped Ordering:] Examples are grouped by task, presenting all demonstrations from one task before moving to the next one (e.g., $t_1, t_1, t_1, t_2, t_2, t_2, t_3, t_3, t_3$). This structure emphasizes intra-task consistency.

\item[Task-Interleaved Ordering:] Examples alternate across tasks at each shot level, maintaining a round-robin pattern (e.g., $t_1, t_2, t_3, t_1, t_2, t_3, t_1, t_2, t_3$). This configuration aims to balance exposure across tasks within the prompt.
\end{description}
The impact of the number of shots per task and example orderings is shown in Appendix~\ref{section:prompts} (Figure~\ref{fig:k_shots} and Figure~\ref{fig:task_ordering}).
Note that to encode examples into dense vector representations, we use the \texttt{jina-embeddings-v3}~\cite{jinaiv3} model, and for distance computation, we employ the L2 distance metric.

\subsection{Multi-Task PEFT}

\paragraph{LoRA Baseline.}
As baseline, we implement LoRA~\cite{hu2022lora}. 
LoRA approximates full fine-tuning by decomposing weight matrices into low-rank components. 
To reduce dimensionality, the weight matrix $\mathbf{W}_0 \in \mathbb{R}^{d \times k}$ is approximated by the product of two lower-rank matrices: $\mathbf{B} \in \mathbb{R}^{d \times r}$ and $\mathbf{A} \in \mathbb{R}^{r \times k}$, with $r \ll \min(d, k)$.
This low-rank update preserves the backbone while enabling efficient adaptation, such that $h = \mathbf{W_0}x + \frac{\alpha}{r}\mathbf{B}\mathbf{A}x$. LoRA can be applied to each linear layer in the Transformer architecture, such as $\mathbf{W_Q}, \mathbf{W_K}, \mathbf{W_V}, \mathbf{W_O}$ matrices projections in the attention layers.


\paragraph{MTL-LoRA.}
\citet{mtllora} introduce MTL-LoRA to face challenges related to task routing and interference mitigation.
Given task input $x_t$, MTL-LoRA first applies a shared standard LoRA down-projection via matrix $\mathbf{A} \in \mathbb{R}^{r \times k}$. To retain task-specific information, it inserts a task-specific low-rank matrix $\Lambda_t \in \mathbb{R}^{r \times r}$ between the down- and up-projections, transforming $\mathbf{A}x_t$. Instead of a single shared up-projection, MTL-LoRA uses $n$ matrices ${\mathbf{B}^i} \in \mathbb{R}^{d \times r}$ to support diverse knowledge-sharing strategies. Outputs are combined via a weighted average, where weights $w_t \in \mathbb{R}^{n \times 1}$ are learned per task as in Equation \ref{eq:mtl}.

\begin{equation}\label{eq:mtl}
\begin{aligned}
h_t = \mathbf{W}x_t + \sum_{i=1}^n \frac{\exp(w_t^i / \tau)\mathbf{B}^i}{\sum_{j=1}^n \exp(w_t^j / \tau)} \Lambda_t \mathbf{A}x_t
\end{aligned}
\end{equation}

\noindent Here, $\tau$ controls the softness of the weighting. Each $\Lambda_t$ is initialized as a diagonal identity matrix to ensure $\Delta\mathbf{W}_t = 0$ at start. 

\paragraph{MTL Loss for ETR Generation.}
The model is trained to generate outputs conditioned on instructions. Given an instruction sequence $I = {i_1, i_2, \ldots, i_m}$ and a corresponding completion sequence $C = {c_1, c_2, \ldots, c_n}$, where $I$ may contain special prompt tokens (e.g., \texttt{<Input>} and \texttt{<Output>}), the full input is represented as $x = {i_1, \ldots, i_m, c_1, \ldots, c_n}$. The model is trained to autoregressively predict each token in $C$ conditioned on all preceding tokens in $I$ and $C$ as defined in Equation \ref{eq:it_prob_emnlp}.

\begin{align}
\label{eq:it_prob_emnlp}
P(C | I) = \prod_{j=1}^n P(c_j \mid i_1,..., i_m, c_1,..., c_{j-1})
\end{align}

\noindent Based on the findings of \citet{huerta-enochian-ko-2024-instruction}, the objective is to minimize the negative log-likelihood of the completion sequence given the instruction as defined in Equation \ref{eq:it_loss}.

\begin{align}
\label{eq:it_loss}
\mathcal{L} = -\sum_{j=1}^n \log P(c_j \mid i_1,..., i_m, c_1,..., c_{j-1})
\end{align}


To account for imbalance across different instruction-following tasks, we apply a task-specific weighting scheme during training. Let $N_t$ be the number of training examples for task $t$, and let $N = \sum_{t} N_{t}$ be the total number of training examples across all tasks. Each task's contribution to the overall loss is scaled by a factor $w_t = \frac{N_t}{N}$, such that the final loss is redefined in Equation \ref{eq:loss_mtl}.

\begin{align}
\label{eq:loss_mtl}
\mathcal{L}_{MTL} = \sum_{t=1}^T w_t \times \mathcal{L}_t
\end{align}

\section{Results}

The top-performing models are chosen according to their highest SRB scores on the \falcdataset validation set, using a grid search strategy for hyperparameter tuning (see Appendix~\ref{apdx: implementation details} for details). 
To complement this analysis, all models are run five times with different seeds, and detailed average results are in Appendix~\ref{app:complementary_results}.

\begin{table*}[ht!]
\small
\centering
\renewcommand{\arraystretch}{1.05}

\subcaptionbox{
\textbf{\falcdataset test set} (In-Domain).\label{tab:ind-results}
}{ 

\resizebox{\textwidth}{!}{
\begin{tabular}{LLLcccccccc}
\toprule
 & \textbf{Method} & \textbf{Task} & \textbf{\RI} & \textbf{\RII} & \textbf{\RL} & \textbf{\sari} & \textbf{\bertf} & \textbf{\SRB} & \textbf{\compratio} & \textbf{\novelty} \\
\midrule
\multicolumn{3}{l}{\textbf{In-Context Learning}} \\ \midrule
\multirow[c]{6}{*}{\rotatebox{90}{\mistVIIinst}} & \zeroshot & E & $23.92$ & $7.09$ & $16.28$ & $37.07$ & $69.75$ & $29.20$ & $-64.14$ & $35.70$ \\
\cmidrule{2-11}
\rotatebox{90}{} & \COT & E & $23.58$ & $7.22$ & $16.17$ & $37.39$ & $68.80$ & $29.10$ & $-60.53$ & $\underline{36.09}$ \\
\cmidrule{2-11}
\rotatebox{90}{} & \multirow[t]{4}{*}{\rag} & E & $32.14$ & $10.47$ & $22.72$ & $40.05$ & $72.41$ & $36.24$ & $44.32$ & $26.55$ \\
\cmidrule{3-11}
\rotatebox{90}{} &  & E,O & $31.12$ & $9.58$ & $21.92$ & $39.54$ & $71.29$ & $35.32$ & $48.45$ & $26.61$ \\
\rotatebox{90}{} &  & E,W & $30.29$ & $9.69$ & $21.29$ & $38.69$ & $71.59$ & $34.56$ & $33.80$ & $23.01$ \\
\rotatebox{90}{} &  & E,O,W & $29.84$ & $9.57$ & $21.58$ & $39.53$ & $71.06$ & $35.01$ & $46.42$ & $25.85$ \\
\cmidrule{1-11}
\multirow[c]{6}{*}{\rotatebox{90}{\llamaVIIIinst}} & \zeroshot & E & $24.94$ & $8.23$ & $17.37$ & $38.59$ & $70.29$ & $30.70$ & $-21.56$ & $\mathbf{38.73}$ \\
\cmidrule{2-11}
\rotatebox{90}{} & \COT & E & $27.57$ & $8.96$ & $18.72$ & $38.26$ & $71.02$ & $32.04$ & $7.80$ & $31.10$ \\
\cmidrule{2-11}
\rotatebox{90}{} & \multirow[t]{4}{*}{\rag} & E & $33.43$ & $12.99$ & $24.38$ & $42.16$ & $72.58$ & $38.21$ & $46.18$ & $27.14$ \\
\cmidrule{3-11}
\rotatebox{90}{} &  & E,O & $31.10$ & $10.87$ & $22.37$ & $39.94$ & $71.27$ & $35.81$ & $39.22$ & $24.29$ \\
\rotatebox{90}{} &  & E,W & $33.03$ & $11.62$ & $23.28$ & $40.59$ & $72.14$ & $36.83$ & $41.89$ & $25.26$ \\
\rotatebox{90}{} &  & E,O,W & $29.35$ & $9.97$ & $20.54$ & $39.03$ & $70.84$ & $33.93$ & $25.94$ & $23.69$ \\
\midrule
\multicolumn{3}{l}{\textbf{Parameter-Efficient Fine-Tuning}} \\  \midrule
\multirow[c]{4}{*}{\rotatebox{90}{\mistVII}} & \lora & E & $32.47$ & $12.40$ & $24.02$ & $42.09$ & $73.56$ & $37.98$ & $44.42$ & $18.35$ \\
\cmidrule{2-11}
\rotatebox{90}{} & \multirow[t]{3}{*}{\mtllora} & E,O & $32.67$ & $12.74$ & $24.33$ & $41.95$ & $73.52$ & $38.20$ & $53.48$ & $24.17$ \\
\rotatebox{90}{} &  & E,W & $32.62$ & $12.92$ & $24.28$ & $42.53$ & $\underline{73.90}$ & $38.35$ & $\underline{53.62}$ & $24.99$ \\
\rotatebox{90}{} &  & E,O,W & $\mathbf{33.65}$ & $12.83$ & $24.93$ & $42.25$ & $73.62$ & $38.77$ & $48.93$ & $23.38$ \\
\cmidrule{1-11}
\multirow[c]{4}{*}{\rotatebox{90}{\llamaVIII}} & \lora & E & $31.76$ & $13.17$ & $25.04$ & $42.15$ & $72.93$ & $38.77$ & $50.66$ & $18.87$ \\
\cmidrule{2-11}
\rotatebox{90}{} & \multirow[t]{3}{*}{\mtllora} & E,O & $\underline{33.44}$ & $13.22$ & $24.24$ & $43.04$ & $73.86$ & $38.45$ & $51.36$ & $23.06$ \\
\rotatebox{90}{} &  & E,W & $32.54$ & $\underline{13.56}$ & $\underline{25.08}$ & $\mathbf{44.67}$ & $\mathbf{74.05}$ & $\underline{39.60}$ & $\mathbf{56.11}$ & $33.05$ \\
\rotatebox{90}{} &  & E,O,W & $32.78$ & $\mathbf{13.64}$ & $\mathbf{25.67}$ & $\underline{43.53}$ & $73.28$ & $\mathbf{39.69}$ & $53.24$ & $24.39$ \\
\bottomrule
\end{tabular}}

 }

\vspace{1em}

\subcaptionbox{
\textbf{\falcdatasetpolitic test set} (Ouf-of-Domain).\label{tab:ood-results}
}{ 
\resizebox{\textwidth}{!}{
\begin{tabular}{LLLcccccccc}
\toprule
 & \textbf{Method} & \textbf{Task} & \textbf{\RI} & \textbf{\RII} & \textbf{\RL} & \textbf{\sari} & \textbf{\bertf} & \textbf{\SRB} & \textbf{\compratio} & \textbf{\novelty} \\
\midrule
\multicolumn{3}{l}{\textbf{In-Context Learning}} \\ \midrule
\multirow[c]{6}{*}{\rotatebox{90}{\mistVIIinst}} & \zeroshot & E & $28.36$ & $11.02$ & $19.29$ & $39.87$ & $68.10$ & $32.75$ & $-309.24$ & $48.37$ \\
\cmidrule{2-11}
\rotatebox{90}{} & \COT & E & $29.78$ & $11.22$ & $19.90$ & $39.62$ & $69.40$ & $33.37$ & $-261.30$ & $\underline{50.85}$ \\
\cmidrule{2-11}
\rotatebox{90}{} & \multirow[t]{4}{*}{\rag} & E & $39.22$ & $15.28$ & $28.12$ & $41.33$ & $73.15$ & $\underline{40.86}$ & $11.03$ & $25.49$ \\
\cmidrule{3-11}
\rotatebox{90}{} &  & E,O & $37.87$ & $14.59$ & $26.43$ & $39.51$ & $72.08$ & $38.96$ & $14.37$ & $18.41$ \\
\rotatebox{90}{} &  & E,W & $39.77$ & $15.55$ & $27.74$ & $40.32$ & $72.47$ & $40.19$ & $10.80$ & $17.81$ \\
\rotatebox{90}{} &  & E,O,W & $39.12$ & $15.97$ & $\underline{28.26}$ & $40.74$ & $72.87$ & $40.73$ & $14.63$ & $18.33$ \\
\cmidrule{1-11}
\multirow[c]{6}{*}{\rotatebox{90}{\llamaVIIIinst}} & \zeroshot & E & $29.60$ & $10.84$ & $18.83$ & $40.55$ & $68.68$ & $32.50$ & $-180.74$ & $\mathbf{55.37}$ \\
\cmidrule{2-11}
\rotatebox{90}{} & \COT & E & $31.68$ & $11.46$ & $20.14$ & $40.80$ & $69.87$ & $33.91$ & $-83.36$ & $45.41$ \\
\cmidrule{2-11}
\rotatebox{90}{} & \multirow[t]{4}{*}{\rag} & E & $37.48$ & $13.98$ & $26.94$ & $41.05$ & $73.18$ & $39.92$ & $11.37$ & $41.63$ \\
\cmidrule{3-11}
\rotatebox{90}{} &  & E,O & $\mathbf{40.53}$ & $15.15$ & $27.47$ & $41.14$ & $72.75$ & $40.29$ & $-12.56$ & $31.01$ \\
\rotatebox{90}{} &  & E,W & $39.72$ & $\underline{16.02}$ & $26.83$ & $\underline{41.99}$ & $\underline{73.32}$ & $40.15$ & $13.75$ & $35.70$ \\
\rotatebox{90}{} &  & E,O,W & $\underline{40.12}$ & $\mathbf{16.55}$ & $\mathbf{28.43}$ & $\mathbf{42.63}$ & $\mathbf{73.39}$ & $\mathbf{41.52}$ & $-4.79$ & $30.08$ \\
\midrule
\multicolumn{3}{l}{\textbf{Parameter-Efficient Fine-Tuning}} \\  \midrule
\multirow[c]{4}{*}{\rotatebox{90}{\mistVII}} & \lora & E & $35.13$ & $12.23$ & $25.93$ & $38.04$ & $70.28$ & $37.94$ & $21.55$ & $11.79$ \\
\cmidrule{2-11}
\rotatebox{90}{} & \multirow[t]{3}{*}{\mtllora} & E,O & $29.36$ & $11.02$ & $21.87$ & $38.68$ & $69.22$ & $34.87$ & $\mathbf{36.68}$ & $40.29$ \\
\rotatebox{90}{} &  & E,W & $34.32$ & $12.56$ & $24.85$ & $38.72$ & $70.54$ & $37.38$ & $\underline{22.51}$ & $19.10$ \\
\rotatebox{90}{} &  & E,O,W & $36.45$ & $13.22$ & $26.21$ & $38.39$ & $70.97$ & $38.32$ & $18.33$ & $10.55$ \\
\cmidrule{1-11}
\multirow[c]{4}{*}{\rotatebox{90}{\llamaVIII}} & \lora & E & $35.53$ & $13.83$ & $26.94$ & $39.90$ & $71.30$ & $39.37$ & $6.38$ & $16.13$ \\
\cmidrule{2-11}
\rotatebox{90}{} & \multirow[t]{3}{*}{\mtllora} & E,O & $32.77$ & $12.20$ & $24.23$ & $38.84$ & $69.74$ & $36.88$ & $18.26$ & $19.30$ \\
\rotatebox{90}{} &  & E,W & $37.46$ & $13.74$ & $27.06$ & $38.26$ & $71.30$ & $38.90$ & $8.45$ & $6.44$ \\
\rotatebox{90}{} &  & E,O,W & $36.48$ & $13.69$ & $25.90$ & $36.19$ & $70.97$ & $37.35$ & $8.68$ & $2.06$ \\
\bottomrule
\end{tabular}}

}
\caption{
\textbf{Performance comparison}, across ICL methods and PEFT strategies on three tasks: \falcdataset (E), \orangesum (O) and \wikilargefr (W). Best results are in \textbf{bold}, second-best are \underline{underlined}.
}
\end{table*}

\subsection{In-Domain Quantitative Results}

\paragraph{ICL Performance.} As shown in Table \ref{tab:ind-results}, ICL models evidence steady improvements when transitioning from zero-shot and CoT prompting to RAG-based prompting. For \llamaVIIIinst, RAG achieves the best results with ETR-fr only inputs (e.g., 33.43/12.99/24.38 ROUGE-1/2/L and 42.16 SARI), outperforming zero-shot by a large margin. Adding related tasks does not consistently improve performance under ICL, and in some cases, leads to reduced novelty and compression ratio.

\paragraph{Impact of Fine-Tuning.} PEFT significantly outperforms ICL methods. The best overall performance is achieved by \llamaVIII with \mtllora fine-tuned on \falcdataset and \wikilargefr, obtaining highest scores across SARI (44.67), BERTScore-F1 (74.05), SRB (39.60), and compression ratio (56.11), while maintaining strong novelty (33.05). 

\paragraph{LLM Comparison.} Across both prompting and fine-tuning paradigms, \llamaVIII outperforms \mistVII in most metrics. For instance, with LoRA fine-tuning on \falcdataset, \llamaVIII achieves higher ROUGE-L (25.04 vs.\ 24.02), SARI (42.15 vs. 42.09), and SRB (38.77 vs. 37.98). This suggests that the architectural or scale advantages of \llamaVIII translate effectively into more efficient capabilities.

\begin{figure*}[ht!]
    \centering
    
    \begin{subfigure}[b]{0.49\textwidth}
        \includegraphics[width=\textwidth]{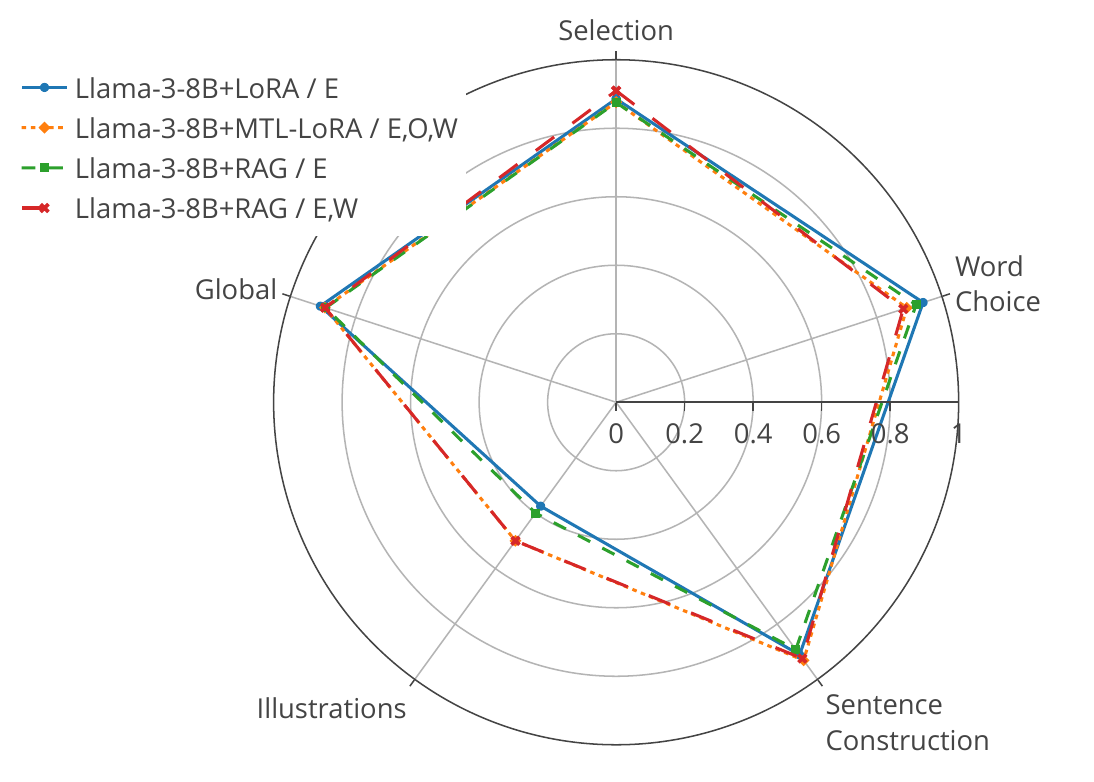}
        \caption{Criteria on \falcdataset}
        \label{fig:valrate_etrfr}
    \end{subfigure}
    \hfill
    \begin{subfigure}[b]{0.49\textwidth}
        \includegraphics[width=\textwidth]{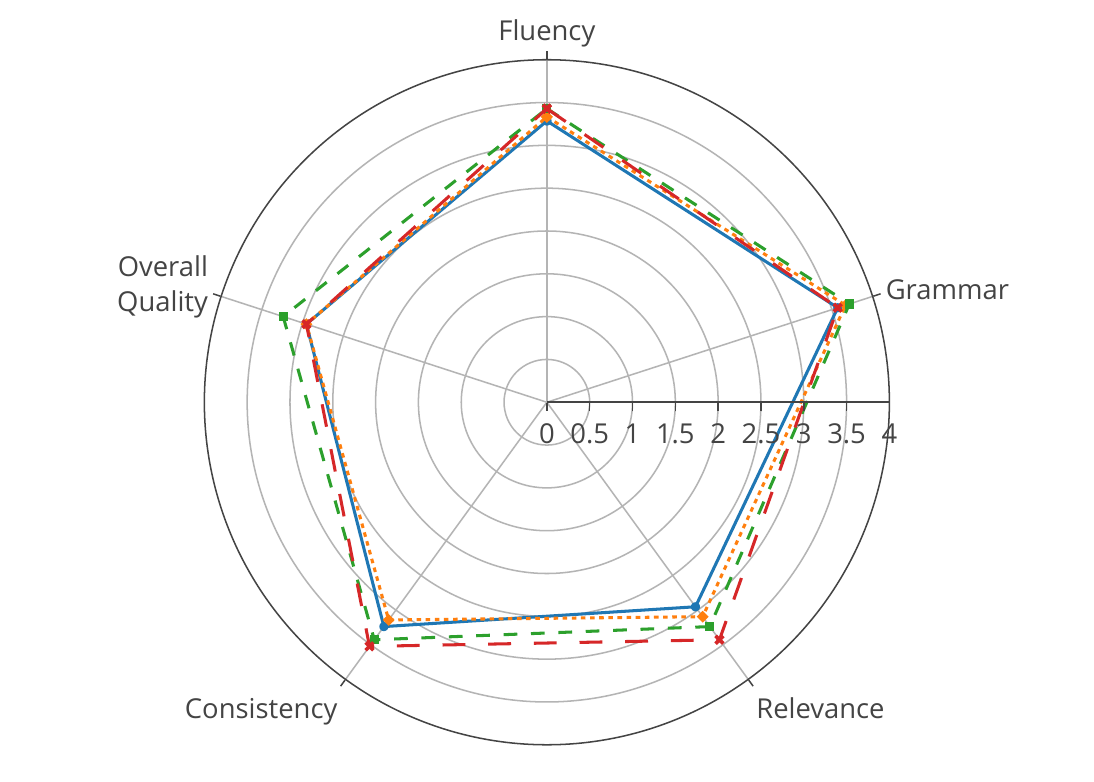}
        \caption{Quality on \falcdataset}
        \label{fig:quality_etrfr}
    \end{subfigure}
    \hfill
    \begin{subfigure}[b]{0.49\textwidth}
        \includegraphics[width=\textwidth]{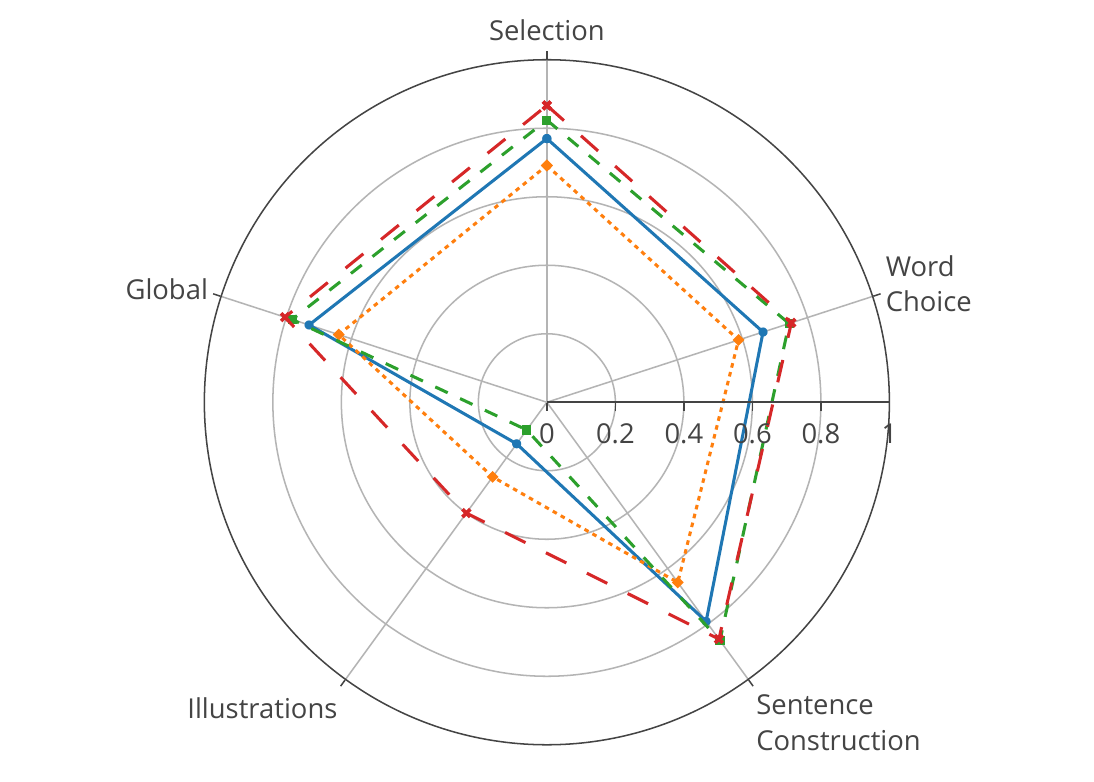}
        \caption{Criteria on \falcdatasetpolitic}
        \label{fig:valrate_etrfrpol}
    \end{subfigure}
    \hfill
    \begin{subfigure}[b]{0.49\textwidth}
        \includegraphics[width=\textwidth]{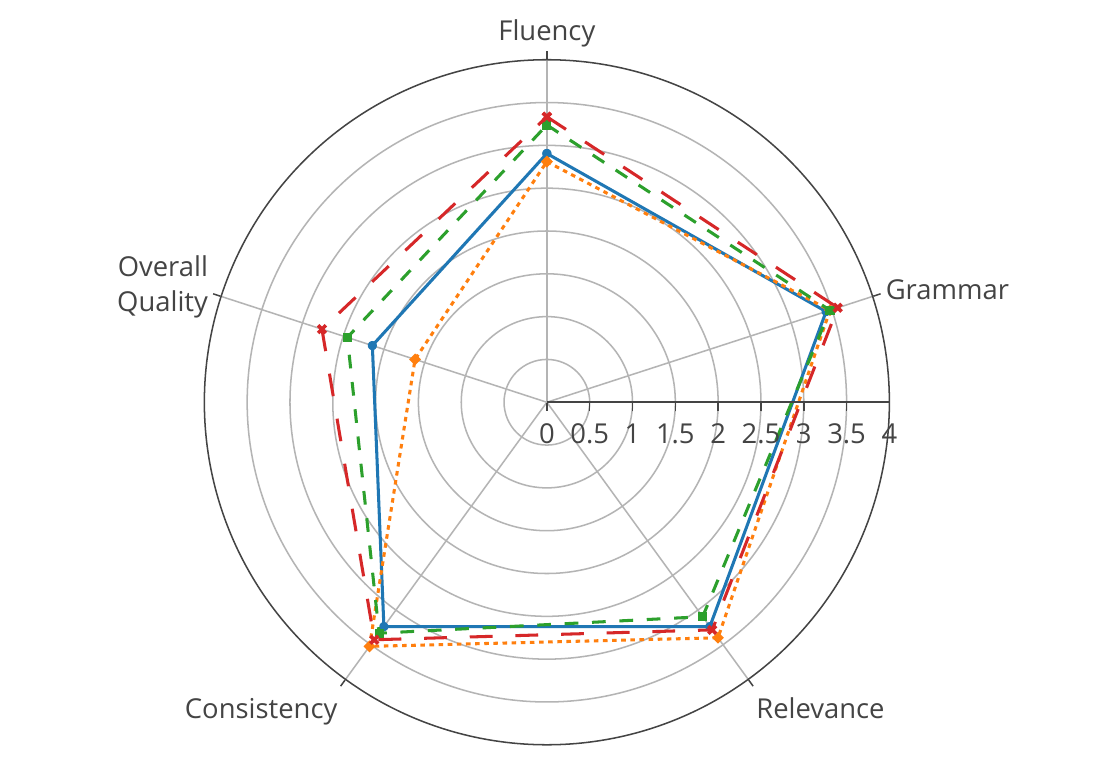}
        \caption{Quality on \falcdatasetpolitic}
        \label{fig:quality_etrfrpol}
    \end{subfigure}

    \caption{
        \textbf{Human evaluation of generation quality on \falcdataset\ and \falcdatasetpolitic} using their optimal ICL and MTL configurations. 
        Subfigures (a) and (c) show average scores based on the ETR guideline criteria. 
        Subfigures (b) and (d) present average human ratings for text generation quality. 
}
    \label{fig:human_eval}
\end{figure*}
\paragraph{Combination of Tasks.} 
Incorporating auxiliary tasks such as text summarization and simplification can provide complementary supervision, as seen in PEFT strategies. However, they do not yield performance gains in the ICL setting. Notably, \mtllora with \falcdataset and \wikilargefr for \llamaVIII achieves the highest SARI and compression ratio, suggesting the relevance of sentence simplification data to the ETR generation task. However, inclusion of all three tasks does not universally yield the best results, and in some cases, introduces performance regressions in BERTScore and novelty. This implies that careful curation of task mixtures is essential to avoid dilution or conflict between training objectives.
Overall, these results highlight that while RAG improves performance in ICL, parameter-efficient fine-tuning (particularly \mtllora) remains the most effective method within the in-domain \falcdataset setting.

\subsection{Out-of-Domain Quantitative Results}

\paragraph{ICL Performance.}
As shown in Table \ref{tab:ood-results}, among prompting strategies, RAG consistently outperforms zero-shot and CoT in all major content preservation metrics (ROUGE-1/2/L, BERTScore-F1) and the composite SRB score. On \llamaVIIIinst, using RAG with all three tasks (E,O,W) achieves the highest overall SRB score (41.52) and the best ROUGE-L (28.43), indicating its strong generalization and content fidelity. Moreover, it yields the highest SARI (42.63) and BERTScore-F1 (73.39), showcasing a balanced ability to simplify while preserving semantics.
Interestingly, zero-shot exhibits extremely poor compression ratios, especially on \mistVIIinst (-309.24), suggesting potential prompt misalignment or excessive hallucination. However, it achieves the highest novelty score (55.37) on \llamaVIIIinst, implying that despite poor content fidelity, more diverse lexical outputs are generated.

\paragraph{Impact of Fine-Tuning.}
While PEFT strategies generally lag behind RAG in terms of SRB and BERTScore, they offer notably better compression ratios than zero-shot, CoT and most RAG-based strategies. The best PEFT model in terms of SRB, LLaMA-3-8B+\lora trained solely on \falcdataset, achieves a relatively low compression ratio (6.38), indicating only moderate summarization. However, this comes at the expense of lower ROUGE, SARI, and BERTScore metrics compared to RAG-based approaches. Additionally, \mtllora configurations do not demonstrate performance improvements over single-task LoRA in out-of-domain (OOD) settings, particularly on \llamaVIII, suggesting a tendency toward overspecialization on the target task of ETR derived from children's books.

\paragraph{Combination of Tasks.}
Prompting or training with multiple datasets (E,O,W) can improve OOD generalization. LLaMA-3-8B+RAG and Mistral-7B+RAG show substantial gains across all metrics compared to single-task prompting, confirming the benefits of multi-domain exposure in OOD settings. This situation is mitigated for the PEFT strategy, where performance improvement is backbone-dependent. While Mistral-7B+MTL-LoRA steadily benefits from concurrent learning achieving best results in terms of SRB with its (E,O,W) configuration, overall best results with LLaMA-3-8B are obtained with a single task setting.  


\subsection{Human Evaluation}
Manual evaluation is essential for assessing ETR text quality and compliance with European guidelines, which include 57 weighted questions covering clarity, simplicity, and accessibility,\footnote{\url{https://www.unapei.org/wp-content/uploads/2020/01/liste_verification-falc-score_v2020-01-14-1.xlsx}} to ensure content is understandable and appropriate for the target audience.
We validated our approach through human evaluation with 10 native French speakers, 7 NLP researchers and 3 linguists, all volunteers, who assessed outputs from the ETR-fr and ETR-politic test sets\footnote{All evaluators received training and were blind to model development to prevent bias.}.
We evaluated outputs generated by two model configurations: (1) Llama-3-8B+RAG augmented with \falcdataset (E) and \wikilargefr (W), and (2) Llama-3-8B+MTL-LoRA trained on \falcdataset, \orangesum (O), and \wikilargefr, alongside their respective single-task variants. These models were chosen as the best performing ones, respectively for ICL and PEFT, for in-domain settings. The evaluation was performed on 6 source documents (3 from \falcdataset and 3 from \falcdatasetpolitic test sets). Each annotator reviewed 24 outputs, resulting in 60 samples per model and a total of 240 different samples evaluated. 
The assessment prioritized the most critical ETR guideline criteria, including information selection, sentence construction, word choice, and illustrations, covering 38 detailed questions (see Table~\ref{tab:questions} in Appendix). Additionally, we assessed general text generation quality metrics such as Fluency, Grammar/Spelling, Relevance, Textual Coherence, and Overall Perceived Quality, through additional five questions. ETR criteria were rated on a binary scale (respected, not respected, not applicable), whereas human judgments used a 5-point Likert scale (0–4).


\paragraph{In-domain Results.}

Figures~\ref{fig:human_eval} presents the human evaluation results and overall scores are provided in a table in Appendix~\ref{sec:appendix_human_evaluation}. On \falcdataset, all methods perform well with respect to the European ETR guidelines. \lora\ achieves the highest overall validation rate of $0.91$, particularly excelling in word choice and sentence construction. MTL-LoRA+(E,O,W) shows the best results for sentence construction, while RAG+(E,W) outperforms other models in information selection.
In terms of text generation quality, single-task RAG leads with an overall score of $4.24$, driven by strong performance in fluency, grammar, and coherence. While MTL-LoRA+(E,O,W) and \lora\ are competitive across individual criteria, with MTL-LoRA+(E,O,W) scoring best on 3 out of 4 dimensions, their overall quality scores are comparable ($3.95$).
Although automatic metrics indicate improved performance in multi-task settings, human evaluation results are more mixed, revealing no clear advantage for single- versus multi-task strategies, except in the Illustrations dimension.


\paragraph{Out-of-domain Results}

Overall performance declines on the more challenging \falcdatasetpolitic, yet RAG+(E,W) remains the most robust across both ETR criteria and text quality evaluations, underscoring the value of the multi-task setting. Specifically, RAG+(E,W), trained on a broader mix of tasks combining ETR and sentence simplification, achieves a total validation rate of $0.80$ for ETR guidelines and an overall quality score of $3.76$. In contrast, MTL-LoRA+(E,O,W) exhibits the sharpest drop in quality ($2.62$), indicating difficulties in managing politically nuanced content, although it still outperforms the single-task configuration in 3 out of 5 evaluation dimensions. Furthermore, in terms of European ETR compliance, MTL-LoRA+(E,O,W) struggles to generalize in out-of-domain settings, showing improvement only in the Illustrations criterion.

\section{Conclusion}

In this paper, we introduced ETR-fr, the first dataset fully compliant with the European ETR guidelines targeting neurodivergent populations, and explored multi-task learning  to improve ETR generation with LLMs. Our experiments show that multi-task setups, particularly RAG for ICL and MTL-LoRA for PEFT, consistently improve performance in both in-domain and OOD settings according to automatic metrics. While human evaluation reveals more nuanced outcomes, it nonetheless confirms the benefits of multi-task learning across a broad range of ETR criteria and text quality dimensions. 
\section{Limitations}

The development of ETR generation models introduces important constraints and considerations that reflect the complexity of cognitive accessibility and language model behavior.



\paragraph{Untested Practical Utility for Users with Disabilities}
While our evaluation combines automatic and human assessments, it does not simulate usage in real-world settings such as assistive reading tools or educational platforms. Thus, the practical utility of outputs for users with intellectual disabilities remains untested.
We aim to support the responsible co-construction of experiments accordingly with ETR inclusion requirements. 
Acknowledging these boundaries also helps position ETR generation as a sociotechnical task, one that demands sensitivity to both linguistic quality and lived experience.


\paragraph{No explicit modeling of cognitive load.}
Though our models optimize for readability and fluency, they do not account for cognitive effort. Even simplified outputs may challenge users when processing abstract or ambiguous content.

\paragraph{ETR guidelines as a fixed supervision target.}
We use European ETR guidelines as a normative framework. While they offer structure, rigid adherence may exclude culturally specific or individualized accessibility strategies, limiting generalization.


\paragraph{Susceptibility to hallucinations.}
As with most generative models, hallucinations and factual drift remain concerns, especially with RAG-based systems. This is particularly risky for audiences who may interpret outputs literally or depend on high textual reliability.

\paragraph{Underexplored ethical considerations.} The automation of content adaptation for cognitive impaired users raises ethical questions around over-simplification, loss of nuance, and possible reinforcement of stereotypes. These dimensions are not addressed in the current evaluation, though they are central to responsible deployment.

\section{Impact and Ethical Considerations}

\paragraph{Risks of Oversimplification.}
Simplified language is not neutral, it involves choices about what meaning is retained or lost. In some cases, simplification may erase nuance, flatten perspective, or reinforce harmful stereotypes. This tension is particularly acute for readers who engage with language differently.

\paragraph{Toward Responsible Design.}
Mitigating risks requires human-in-the-loop systems, participatory evaluation involving end users, and adaptation strategies that go beyond surface-level clarity. ETR guidelines should be viewed as a starting point, not a universal solution.

\paragraph{Positioning ETR as a Research Problem.}
ETR remains underexplored in NLP. By introducing aligned data, task-specific metrics, and a critical lens on modeling assumptions, we aim to establish it as a standalone task, one that demands linguistic sensitivity, practical design, and participatory validation.

\section*{Acknowledgments}

We express our gratitude to François Baudez Publishing for generously granting free access to the \falc book collection, which enabled the construction of our dataset. 

We also acknowledge the annotators for their thorough and precise contributions to the manual evaluation, which were instrumental in ensuring the validity and robustness of our results. 

Finally, this work was carried out using the computing resources provided by CRIANN (Normandy, France).

\bibliography{etr_mtl}

\appendix

\section{Implementation Details}
\label{apdx: implementation details}


\subsection{Multi-Task Methods}
\paragraph{Finetuning.}
LLMs are trained for 6 epochs maximum, using the AdamW optimizer~\cite{loshchilov_decoupled_2019} with the following parameters: $\epsilon=10^{-9}$, $\beta_1=0.9$, $\beta_2=0.999$, and a weight decay of $\lambda=0.01$. 
A linear learning rate scheduler with a 10\% warm-up ratio is employed. 
The training batch size is fixed at 4, with 4 steps gradient accumulation and training tasks are randomly sampled.
The learning rate is chosen from the set $\{1\cdot10^{-5}, 2\cdot10^{-5}, 5\cdot10^{-5}, 1\cdot10^{-4}\}$, and hyperparameter selection is performed to maximize SRB (see \S \ref{apdx:SRB}).
According to experimental findings, \lora and \mtllora hyperparameters are set to $r=128$ and $attn\_matrices = \Wqkvo$. Moreover, we chose $\alpha = r$ to keep a 1:1 ratio so as not to overpower the backbone \cite{lee_platypus_2023}.
For \mtllora configuration, sharpness of the weight distribution is fixed at $0.5$ and the optimal $n$ up-projections is selected among $\{1, 2, 3\}$.
We rely on the implementation provided by \texttt{Adapters} library \cite{poth_adapters_2023} for all PEFT methods.
Best hyperparameters for PEFT methods are in Table~\ref{tab:hps_peft}.

\paragraph{MTL-RAG.}
\begin{table*}[ht!]
\centering
\tiny
\resizebox{\textwidth}{!}{\begin{tabular}{LLLcccccccc}
\toprule
 &  &  & \textbf{Batch size} & \textbf{lr} & \textbf{Acc. steps} & \textbf{Epochs} & \textbf{$\alpha = r$} & \textbf{Attn. matrices} & \textbf{n up proj.} & \textbf{$\tau$} \\
\midrule
\multirow[c]{4}{*}{\rotatebox{90}{\llamaVIII}} & \lora & E & 4 & $1\cdot10^{-4}$ & 4 & 6 & 128 & $W_{QKVO}$ & - & - \\
\cmidrule{2-11}
\rotatebox{90}{} & \multirow[t]{3}{*}{\mtllora} & E,O,W & 4 & $1\cdot10^{-4}$ & 4 & 6 & 128 & $W_{QKVO}$ & 3 & 0.5 \\
\rotatebox{90}{} &  & E,O & 4 & $1\cdot10^{-4}$ & 4 & 6 & 128 & $W_{QKVO}$ & 3 & 0.5 \\
\rotatebox{90}{} &  & E,W & 4 & $1\cdot10^{-4}$ & 4 & 6 & 128 & $W_{QKVO}$ & 3 & 0.5 \\
\cmidrule{1-11} \cmidrule{2-11}
\multirow[c]{4}{*}{\rotatebox{90}{\mistVII}} & \lora & E & 4 & $1\cdot10^{-4}$ & 4 & 6 & 128 & $W_{QKVO}$ & - & - \\
\cmidrule{2-11}
\rotatebox{90}{} & \multirow[t]{3}{*}{\mtllora} & E,O,W & 4 & $1\cdot10^{-4}$ & 4 & 6 & 128 & $W_{QKVO}$ & 3 & 0.5 \\
\rotatebox{90}{} &  & E,O & 4 & $5\cdot10^{-5}$ & 4 & 6 & 128 & $W_{QKVO}$ & 3 & 0.5 \\
\rotatebox{90}{} &  & E,W & 4 & $1\cdot10^{-4}$ & 4 & 6 & 128 & $W_{QKVO}$ & 3 & 0.5 \\

\bottomrule
\end{tabular}}
\caption{PEFT hyperparameter configurations chosen based on SRB performance on the \falcdataset validation set. E, O, and W refer to \falcdataset, \orangesum, and \wikilargefr, respectively.}
\label{tab:hps_peft}
\end{table*}
\begin{table}
\centering
\small
\begin{tabular}{LLLcc}
\toprule
 &  &  & $\mathbf{k}$ & \textbf{Ordering} \\
\midrule
\multirow[c]{6}{*}{\rotatebox{90}{\mistVIIinst}} & \zeroshot & E & - & - \\
\cmidrule{2-5}
\rotatebox{90}{} & \COT & E & - & - \\
\cmidrule{2-5}
\rotatebox{90}{} & \multirow[t]{4}{*}{\rag} & E & 7 & Random \\
\cmidrule{3-5}
\rotatebox{90}{} &  & E,O & 3 & Random \\
\rotatebox{90}{} &  & E,W & 3 & Random \\
\rotatebox{90}{} &  & E,O,W & 3 & Interleaved \\
\cmidrule{1-5}
\multirow[c]{6}{*}{\rotatebox{90}{\llamaVIIIinst}} & \zeroshot & E & - & - \\
\cmidrule{2-5}
\rotatebox{90}{} & \COT & E & - & - \\
\cmidrule{2-5}
\rotatebox{90}{} & \multirow[t]{4}{*}{\rag} & E & 9 & Random \\
\cmidrule{3-5}
\rotatebox{90}{} &  & E,O & 3 & Random \\
\rotatebox{90}{} &  & E,W & 3 & Random \\
\rotatebox{90}{} &  & E,O,W & 2 & Random \\

\bottomrule
\end{tabular}
\caption{ICL hyperparameter configurations selected based on SRB performance on the \falcdataset validation set. Here, E denotes \falcdataset, O denotes \orangesum, and W denotes \wikilargefr.}
\label{tab:hps_icl}
\end{table}

To facilitate few-shot multi-task learning within the in-context learning framework, we develop a multi-task extension of Retrieval-Augmented Generation (RAG). Our approach retrieves demonstrations from various tasks and integrates them into the prompt. We conduct experiments using 1, 2, and 3 examples per task, analyzing how the ordering of tasks and examples within the prompt influences the performance.
We investigate three strategies for sequencing demonstrations in the prompt as mentioned in Section~\ref{sec:mtl_rag}: random, grouped and interleaved orderings. 
Detailled results are in Appendix~\ref{section:prompts}.

The optimal hyperparameters for in-context learning are summarized in Table~\ref{tab:hps_icl}.

\subsection{Models}

We utilize the following instruct models for In-Context Learning (ICL):
\begin{itemize}
    \item \href{https://huggingface.co/meta-llama/Llama-3.1-8B-Instruct}{\texttt{Llama-3.1-8B-Instruct}}
    \item \href{https://huggingface.co/mistralai/Mistral-7B-Instruct-v0.3}{\texttt{Mistral-7B-Instruct-v0.3}}
\end{itemize}

For experiments involving Parameter-Efficient Fine-Tuning (PEFT), we employ the following base models:
\begin{itemize}
    \item \href{https://huggingface.co/meta-llama/Llama-3.1-8B}{\texttt{Llama-3.1-8B}}
    \item \href{https://huggingface.co/mistralai/Mistral-7B-v0.3}{\texttt{Mistral-7B-v0.3}}
    \item \href{https://huggingface.co/deepseek-ai/DeepSeek-R1-Distill-Llama-8B}{\texttt{DeepSeek-R1-Distill-Llama-8B}}
\end{itemize}

\subsection{Metrics}\label{apdx: metrics}

\paragraph{Text Descriptive Statistics.}

To compute the descriptive statistics presented in Table~\ref{tab:dataset_comparaison}, such as word count, sentence length, compression ratio, KMRE, and others, we employ the \href{https://github.com/HLasse/TextDescriptives}{\texttt{TextDescriptives}} \cite{Hansen_2023} and \href{https://github.com/chartbeat-labs/textacy}{\texttt{textacy}} Python libraries, both of which use the \texttt{fr\_core\_news\_md-3.8.0} model from \href{https://spacy.io}{\texttt{SpaCy}}.

\paragraph{\rouge} (Recall-Oriented Understudy for Gisting Evaluation) \cite{lin_rouge_2004} is a widely used metric for assessing the quality of automatically generated summaries by measuring n-gram and sequence overlap with reference texts. Specifically, we report the F1-scores for \rone (ROUGE-1), \rtwo (ROUGE-2), and \rl (ROUGE-L), which capture overlap of unigrams, bigrams, and longest common subsequences, respectively. The F1-score represents the harmonic mean of precision and recall. 
For evaluation, we use \href{https://huggingface.co/spaces/evaluate-metric/rouge}{Hugging Face's interface} to \href{https://github.com/google-research/google-research/tree/master/rouge}{Google’s official implementation}.

\paragraph{\bertscore} \cite{zhang_bertscore_2020} is based on the contextual word representations generated by BERT-like encoders. Unlike traditional metrics like \bleu or \rouge, which rely on exact lexical matches, \bertscore uses embeddings to capture finer semantic similarities, offering more flexibility with respect to context and greater robustness to word reordering and synonyms. For each word in the generated text, BERTScore finds the most similar word in the reference text using cosine similarities of their representations. The goal of this step is to align the words in the generated text with those in the reference text. These similarity scores for the aligned word pairs are then aggregated to obtain recall, precision, and \fscore. 
For reproducibility, we use the \href{https://huggingface.co/spaces/evaluate-metric/bertscore}{Hugging Face's wrapper} coupled with \texttt{bert-base-multilingual-cased} model.


\paragraph{SARI} (Sentence-level Accuracy Rating for Text Simplification) \cite{xu_optimizing_2016} is commonly used to evaluate sentence and text simplification. Unlike other metrics like \bleu or \rouge, which focus primarily on lexical similarity to reference texts, SARI takes into account three key aspects of simplification: content preservation (\texttt{keep}), information addition (\texttt{add}), and information deletion (\texttt{del}). For each word or n-gram generated, SARI evaluates whether the word should be kept, added, or deleted by comparing it with its source and the ground truth. The mathematical expression of SARI is the average of the \fscore of these three measures.

\begin{equation*}
    \text{SARI} = \frac{F1_{\texttt{keep}} + F1_{\texttt{add}} + F1_{\texttt{del}}}{3}
\end{equation*}

For evaluation, we use \href{https://huggingface.co/spaces/evaluate-metric/sari}{Hugging Face's interface}, which is adapted from TensorFlow’s \texttt{tensor2tensor} implementation \cite{tensor2tensor}.

\paragraph{KMRE}\label{sec:kmre} (Kandel-Moles Reading Ease) \cite{kandel_application_1958} is the French adaptation of the Flesch-Kincaid Reading Ease (FKRE) \cite{kincaid_derivation_1975}, originally designed for English. It measures the complexity of French texts based on sentence length and word length without the need for comparison with a reference text:  

\begin{equation*}
\scriptsize
    \text{\kmre} = \text{207 - 1.015} \left( \dfrac{\text{\#words}}{\text{\#sentences}} \right)
    - \text{73.6} \left( \dfrac{\text{\#syllables}}{\text{\#words}} \right) 
\end{equation*}

KMRE, like the FKRE, is theoretically bounded between 0 and 100. However, it can exceed 100 in rare cases, particularly when the text contains very short sentences and simple, monosyllabic words. This is often the case in ETR documents, which are specifically designed for ease of reading.
Moreover, \citet{wubben_sentence_2012} advises not to use this metric alone, as it does not account for grammar quality or meaning preservation. This is why we pair it with \bertscore, \rouge, and SARI, and we do not monitor it for hyperparameter tuning.



\paragraph{SRB}\label{apdx:SRB} is proposed to measure the quality of a \falc text by aggregating metrics related to \falc transcription characteristics, \ie simplification, summarization, and meaning preservation. To do this, we compute the harmonic mean of SARI, \rl, and \bertscoref: 
\begin{equation*}
    \text{SRB} = \dfrac{3}
    {\dfrac{1}{\text{SARI}} + \dfrac{1}{\text{R-L}} + \dfrac{1}{\text{\bertscoref}}}    
\end{equation*}

\paragraph{Novelty} is used to evaluate abstractiveness, measured by the percentage of n-grams in the generated text that do not appear in the source document \cite{see-etal-2017-get, kamal_eddine_barthez_2021}. We report only novel 1-grams, excluding stop words (commonly used words in a language).

\paragraph{Compression ratio} is the proportion of the document that has been removed. A higher compression ratio indicates more reduction, meaning the summary is more condensed compared to the original document.

\begin{equation*}
    \text{Comp. Ratio} = 1 - \dfrac{\text{\#words in \falc}}{\text{\#words in source}}
\end{equation*}
\section{Complementary Evaluation Results}
\label{app:complementary_results}

\subsection{Quantitative Results}
\label{apdx: qtt results}
Average performances of various methods on \falcdataset and \falcdatasetpolitic test sets are presented in tables \ref{tab:avg-results} and \ref{tab:ood-avg-results}, respectively. These results compare In-Context Learning (ICL) techniques, such as Zero-shot, Chain-of-Thought (CoT), and Retrieval-Augmented Generation (RAG), against Parameter-Efficient Fine-Tuning (PEFT) methods including LoRA and MTL-LoRA. Evaluations are conducted across different LLM models (\mistVII, \llamaVIII and \deepseekrI) and task combinations (E: \falcdataset, O: \orangesum, W: \wikilargefr). Metrics such as ROUGE (R-1, R-2, R-L), SARI, BERTScore-F1, SRB, Compression Ratio, and Novelty are used to provide a comprehensive performance overview.

\newcommand{\rot}[2][90]{\rotatebox{#1}{\parbox{1.5cm}{\centering #2}}}
\begin{table*}[!ht]

\small
\centering

\renewcommand{\arraystretch}{1}

    \subcaptionbox{
{Performance on \falcdataset test set}.\label{tab:avg-results}
}{
    \adjustbox{max width=\textwidth}{
\begin{tabular}{LLLcccccccc}
\toprule
 & \textbf{Method} & \textbf{Task} & \textbf{\RI} & \textbf{\RII} & \textbf{\RL} & \textbf{\sari} & \textbf{\bertf} & \textbf{\SRB} & \textbf{\compratio} & \textbf{\novelty} \\
\midrule
\multicolumn{3}{c}{\textbf{In Context Learning}} \\ \midrule
\multirow[c]{6}{*}{\rot{\mistVIIinst}} & \zeroshot & E & $23.96_{\pm0.04}$ & $7.08_{\pm0.01}$ & $16.25_{\pm0.03}$ & $37.07_{\pm0.00}$ & $69.75_{\pm0.00}$ & $29.17_{\pm0.03}$ & $-64.14_{\pm0.00}$ & $35.70_{\pm0.00}$ \\
\cmidrule{2-11}
\rot{} & \COT & E & $23.53_{\pm0.06}$ & $7.23_{\pm0.01}$ & $16.20_{\pm0.04}$ & $37.39_{\pm0.00}$ & $68.80_{\pm0.00}$ & $29.12_{\pm0.05}$ & $-60.53_{\pm0.00}$ & $\underline{36.09}_{\pm0.00}$ \\
\cmidrule{2-11}
\rot{} & \multirow[t]{4}{*}{\rag} & E & $\underline{31.91}_{\pm0.66}$ & $\underline{10.77}_{\pm0.65}$ & $\underline{22.54}_{\pm0.75}$ & $\underline{40.14}_{\pm0.57}$ & $\underline{72.17}_{\pm0.30}$ & $\underline{36.08}_{\pm0.80}$ & $45.23_{\pm1.17}$ & $27.27_{\pm0.58}$ \\
\cmidrule{3-11}
\rot{} &  & E,O & $30.36_{\pm0.47}$ & $9.61_{\pm0.34}$ & $21.80_{\pm0.30}$ & $39.49_{\pm0.12}$ & $71.07_{\pm0.18}$ & $35.19_{\pm0.29}$ & $\underline{47.99}_{\pm1.91}$ & $26.80_{\pm0.84}$ \\
\rot{} &  & E,W & $30.46_{\pm0.48}$ & $9.93_{\pm0.17}$ & $21.72_{\pm0.34}$ & $38.76_{\pm0.43}$ & $71.57_{\pm0.14}$ & $34.96_{\pm0.34}$ & $35.08_{\pm2.13}$ & $23.32_{\pm0.31}$ \\
\rot{} &  & E,O,W & $29.85_{\pm0.04}$ & $9.58_{\pm0.03}$ & $21.55_{\pm0.05}$ & $39.53_{\pm0.00}$ & $71.06_{\pm0.00}$ & $34.98_{\pm0.05}$ & $46.42_{\pm0.00}$ & $25.85_{\pm0.00}$ \\
\cmidrule{1-11}
\multirow[c]{6}{*}{\rot{\llamaVIIIinst}} & \zeroshot & E & $24.90_{\pm0.20}$ & $8.16_{\pm0.25}$ & $17.10_{\pm0.38}$ & $38.48_{\pm0.38}$ & $70.15_{\pm0.17}$ & $30.38_{\pm0.48}$ & $-22.52_{\pm2.47}$ & $\mathbf{39.13}_{\pm0.92}$ \\
\cmidrule{2-11}
\rot{} & \COT & E & $27.23_{\pm0.91}$ & $8.81_{\pm0.21}$ & $18.34_{\pm0.57}$ & $38.15_{\pm0.23}$ & $70.79_{\pm0.52}$ & $31.62_{\pm0.65}$ & $7.59_{\pm4.82}$ & $30.33_{\pm1.75}$ \\
\cmidrule{2-11}
\rot{} & \multirow[t]{4}{*}{\rag} & E & $\underline{33.05}_{\pm0.72}$ & $\underline{12.23}_{\pm0.44}$ & $\underline{23.77}_{\pm0.68}$ & $\underline{41.66}_{\pm0.45}$ & $\underline{72.59}_{\pm0.38}$ & $\underline{37.57}_{\pm0.70}$ & $\underline{43.36}_{\pm2.62}$ & $27.06_{\pm0.29}$ \\
\cmidrule{3-11}
\rot{} &  & E,O & $30.77_{\pm0.35}$ & $10.85_{\pm0.31}$ & $22.10_{\pm0.35}$ & $39.84_{\pm0.22}$ & $71.13_{\pm0.17}$ & $35.54_{\pm0.32}$ & $24.36_{\pm30.13}$ & $25.02_{\pm1.84}$ \\
\rot{} &  & E,W & $32.14_{\pm0.56}$ & $11.70_{\pm0.34}$ & $23.11_{\pm0.19}$ & $40.49_{\pm0.32}$ & $71.88_{\pm0.18}$ & $36.64_{\pm0.24}$ & $42.30_{\pm1.59}$ & $26.70_{\pm0.92}$ \\
\rot{} &  & E,O,W & $30.53_{\pm0.74}$ & $10.67_{\pm0.45}$ & $21.65_{\pm0.71}$ & $39.24_{\pm0.20}$ & $71.21_{\pm0.26}$ & $35.00_{\pm0.67}$ & $31.18_{\pm4.94}$ & $24.08_{\pm1.37}$ \\
\cmidrule{1-11}
\multicolumn{3}{c}{\textbf{PEFT}} \\  \midrule
\multirow[c]{4}{*}{\rot{\mistVII}} & \lora & E & $32.45_{\pm0.03}$ & $12.38_{\pm0.02}$ & $23.99_{\pm0.05}$ & $42.09_{\pm0.00}$ & $73.56_{\pm0.00}$ & $37.95_{\pm0.04}$ & $44.42_{\pm0.00}$ & $18.35_{\pm0.00}$ \\
\cmidrule{2-11}
\rot{} & \multirow[t]{3}{*}{\mtllora} & E,O & $32.62_{\pm0.04}$ & $12.73_{\pm0.01}$ & $24.29_{\pm0.04}$ & $41.95_{\pm0.00}$ & $73.52_{\pm0.00}$ & $38.16_{\pm0.03}$ & $53.48_{\pm0.00}$ & $24.17_{\pm0.00}$ \\
\rot{} &  & E,W & $32.68_{\pm0.05}$ & $\underline{12.91}_{\pm0.01}$ & $24.25_{\pm0.03}$ & $\underline{42.53}_{\pm0.00}$ & $\underline{73.90}_{\pm0.00}$ & $38.33_{\pm0.03}$ & $\underline{53.62}_{\pm0.00}$ & $\underline{24.99}_{\pm0.00}$ \\
\rot{} &  & E,O,W & $\mathbf{33.60}_{\pm0.05}$ & $12.81_{\pm0.05}$ & $\underline{24.89}_{\pm0.04}$ & $42.25_{\pm0.00}$ & $73.62_{\pm0.00}$ & $\underline{38.74}_{\pm0.03}$ & $48.93_{\pm0.00}$ & $23.38_{\pm0.00}$ \\
\cmidrule{1-11}
\multirow[c]{4}{*}{\rot{\llamaVIII}} & \lora & E & $31.80_{\pm0.03}$ & $13.16_{\pm0.09}$ & $24.92_{\pm0.18}$ & $42.15_{\pm0.01}$ & $72.84_{\pm0.17}$ & $38.67_{\pm0.17}$ & $50.50_{\pm0.28}$ & $18.37_{\pm0.88}$ \\
\cmidrule{2-11}
\rot{} & \multirow[t]{3}{*}{\mtllora} & E,O & $\underline{33.38}_{\pm0.06}$ & $13.16_{\pm0.05}$ & $24.20_{\pm0.04}$ & $43.06_{\pm0.01}$ & $73.88_{\pm0.01}$ & $38.42_{\pm0.03}$ & $50.90_{\pm0.40}$ & $23.25_{\pm0.17}$ \\
\rot{} &  & E,W & $32.54_{\pm0.05}$ & $13.50_{\pm0.06}$ & $25.01_{\pm0.06}$ & $\mathbf{44.67}_{\pm0.00}$ & $\mathbf{74.05}_{\pm0.00}$ & $39.54_{\pm0.05}$ & $\underline{56.11}_{\pm0.00}$ & $\underline{33.05}_{\pm0.00}$ \\
\rot{} &  & E,O,W & $32.78_{\pm0.02}$ & $\mathbf{13.67}_{\pm0.03}$ & $\mathbf{25.55}_{\pm0.16}$ & $43.58_{\pm0.10}$ & $73.33_{\pm0.09}$ & $\mathbf{39.62}_{\pm0.09}$ & $52.66_{\pm1.00}$ & $24.27_{\pm0.21}$ \\
\cmidrule{1-11}
\multirow[c]{4}{*}{\rot{\deepseekrI}} & \lora & E & $20.45_{\pm0.65}$ & $7.72_{\pm0.29}$ & $15.40_{\pm0.13}$ & $41.29_{\pm0.04}$ & $66.02_{\pm0.26}$ & $28.76_{\pm0.16}$ & $-4.61_{\pm3.83}$ & $21.86_{\pm0.29}$ \\
\cmidrule{2-11}
\rot{} & \multirow[t]{3}{*}{\mtllora} & E,O & $23.70_{\pm0.32}$ & $8.86_{\pm0.04}$ & $18.18_{\pm0.33}$ & $42.91_{\pm0.06}$ & $66.72_{\pm0.24}$ & $32.15_{\pm0.37}$ & $\underline{8.57}_{\pm1.08}$ & $27.92_{\pm0.86}$ \\
\rot{} &  & E,W & $\underline{25.38}_{\pm0.11}$ & $\underline{9.35}_{\pm0.05}$ & $\underline{18.52}_{\pm0.07}$ & $\underline{43.06}_{\pm0.03}$ & $\underline{68.08}_{\pm0.14}$ & $\underline{32.64}_{\pm0.08}$ & $-0.52_{\pm2.52}$ & $\underline{36.16}_{\pm0.30}$ \\
\rot{} &  & E,O,W & $22.70_{\pm0.10}$ & $7.93_{\pm0.01}$ & $16.59_{\pm0.02}$ & $42.94_{\pm0.00}$ & $67.18_{\pm0.00}$ & $30.47_{\pm0.02}$ & $-9.35_{\pm0.00}$ & $29.50_{\pm0.00}$ \\
\bottomrule
\end{tabular}
}}
\vspace{1em}

\subcaptionbox{
{Performance on \falcdatasetpolitic test set}.\label{tab:ood-avg-results}
}{
    \adjustbox{max width=\textwidth}{

\begin{tabular}{LLLcccccccc}
\toprule
 & \textbf{Method} & \textbf{Task} & \textbf{\RI} & \textbf{\RII} & \textbf{\RL} & \textbf{\sari} & \textbf{\bertf} & \textbf{\SRB} & \textbf{\compratio} & \textbf{\novelty} \\
\midrule
\multicolumn{3}{c}{\textbf{In Context Learning}} \\ \midrule
\multirow[c]{6}{*}{\rot{\mistVIIinst}} & \zeroshot & E & $28.42_{\pm0.12}$ & $10.98_{\pm0.07}$ & $19.31_{\pm0.03}$ & $39.87_{\pm0.00}$ & $68.10_{\pm0.00}$ & $32.77_{\pm0.03}$ & $-309.24_{\pm0.00}$ & $48.37_{\pm0.00}$ \\
\cmidrule{2-11}
\rot{} & \COT & E & $29.80_{\pm0.03}$ & $11.21_{\pm0.05}$ & $19.88_{\pm0.08}$ & $39.62_{\pm0.00}$ & $69.40_{\pm0.00}$ & $33.35_{\pm0.07}$ & $-261.30_{\pm0.00}$ & $\underline{50.85}_{\pm0.00}$ \\
\cmidrule{2-11}
\rot{} & \multirow[t]{4}{*}{\rag} & E & $\mathbf{40.19}_{\pm0.63}$ & $\underline{16.07}_{\pm0.60}$ & $28.25_{\pm0.31}$ & $\underline{41.40}_{\pm0.46}$ & $\underline{73.01}_{\pm0.34}$ & $\underline{40.96}_{\pm0.35}$ & $9.00_{\pm3.96}$ & $23.21_{\pm2.39}$ \\
\cmidrule{3-11}
\rot{} &  & E,O & $37.49_{\pm0.61}$ & $14.50_{\pm0.35}$ & $26.38_{\pm0.69}$ & $39.46_{\pm0.35}$ & $72.27_{\pm0.26}$ & $38.92_{\pm0.58}$ & $14.26_{\pm2.65}$ & $17.57_{\pm1.61}$ \\
\rot{} &  & E,W & $39.65_{\pm0.19}$ & $15.36_{\pm0.35}$ & $27.85_{\pm0.38}$ & $40.08_{\pm0.36}$ & $72.35_{\pm0.29}$ & $40.17_{\pm0.23}$ & $8.72_{\pm1.73}$ & $17.47_{\pm1.68}$ \\
\rot{} &  & E,O,W & $39.14_{\pm0.04}$ & $15.96_{\pm0.09}$ & $\underline{28.40}_{\pm0.11}$ & $40.74_{\pm0.00}$ & $72.87_{\pm0.00}$ & $40.82_{\pm0.07}$ & $\underline{14.63}_{\pm0.00}$ & $18.33_{\pm0.00}$ \\
\cmidrule{1-11}
\multirow[c]{6}{*}{\rot{\llamaVIIIinst}} & \zeroshot & E & $29.10_{\pm0.40}$ & $10.68_{\pm0.35}$ & $18.70_{\pm0.41}$ & $40.68_{\pm0.48}$ & $68.65_{\pm0.11}$ & $32.39_{\pm0.51}$ & $-178.23_{\pm7.77}$ & $\mathbf{55.73}_{\pm1.07}$ \\
\cmidrule{2-11}
\rot{} & \COT & E & $31.15_{\pm0.99}$ & $10.47_{\pm0.81}$ & $19.54_{\pm0.65}$ & $39.80_{\pm0.63}$ & $69.66_{\pm0.43}$ & $33.09_{\pm0.74}$ & $-70.57_{\pm8.09}$ & $47.80_{\pm1.71}$ \\
\cmidrule{2-11}
\rot{} & \multirow[t]{4}{*}{\rag} & E & $37.68_{\pm0.53}$ & $14.46_{\pm0.65}$ & $26.09_{\pm0.60}$ & $42.05_{\pm0.90}$ & $73.01_{\pm0.20}$ & $39.57_{\pm0.41}$ & $1.47_{\pm6.45}$ & $41.78_{\pm0.86}$ \\
\cmidrule{3-11}
\rot{} &  & E,O & $37.43_{\pm2.11}$ & $14.28_{\pm0.89}$ & $25.92_{\pm1.42}$ & $40.95_{\pm0.90}$ & $72.41_{\pm0.61}$ & $39.05_{\pm1.37}$ & $-7.72_{\pm14.32}$ & $31.85_{\pm1.69}$ \\
\rot{} &  & E,W & $\underline{39.99}_{\pm1.10}$ & $\mathbf{16.27}_{\pm0.61}$ & $\underline{27.84}_{\pm1.10}$ & $\mathbf{42.41}_{\pm0.43}$ & $\mathbf{73.83}_{\pm0.47}$ & $\mathbf{41.06}_{\pm0.96}$ & $\underline{13.46}_{\pm2.37}$ & $36.72_{\pm2.01}$ \\
\rot{} &  & E,O,W & $38.33_{\pm1.46}$ & $15.12_{\pm1.08}$ & $26.89_{\pm1.10}$ & $41.08_{\pm0.94}$ & $72.86_{\pm0.51}$ & $39.86_{\pm1.13}$ & $6.34_{\pm7.54}$ & $29.92_{\pm0.48}$ \\
\cmidrule{1-11}
\multicolumn{3}{c}{\textbf{PEFT}} \\  \midrule
\multirow[c]{4}{*}{\rot{\mistVII}} & \lora & E & $35.10_{\pm0.04}$ & $12.28_{\pm0.04}$ & $25.97_{\pm0.03}$ & $38.04_{\pm0.00}$ & $70.28_{\pm0.00}$ & $37.96_{\pm0.02}$ & $21.55_{\pm0.00}$ & $11.79_{\pm0.00}$ \\
\cmidrule{2-11}
\rot{} & \multirow[t]{3}{*}{\mtllora} & E,O & $29.29_{\pm0.07}$ & $11.02_{\pm0.01}$ & $21.90_{\pm0.04}$ & $38.68_{\pm0.00}$ & $69.22_{\pm0.00}$ & $34.90_{\pm0.03}$ & $\mathbf{36.68}_{\pm0.00}$ & $\underline{40.29}_{\pm0.00}$ \\
\rot{} &  & E,W & $34.32_{\pm0.06}$ & $12.60_{\pm0.07}$ & $24.87_{\pm0.11}$ & $\underline{38.72}_{\pm0.00}$ & $70.54_{\pm0.00}$ & $37.40_{\pm0.09}$ & $22.51_{\pm0.00}$ & $19.10_{\pm0.00}$ \\
\rot{} &  & E,O,W & $\underline{36.34}_{\pm0.10}$ & $\underline{13.24}_{\pm0.02}$ & $\underline{26.29}_{\pm0.08}$ & $38.39_{\pm0.00}$ & $\underline{70.97}_{\pm0.00}$ & $\underline{38.37}_{\pm0.06}$ & $18.33_{\pm0.00}$ & $10.55_{\pm0.00}$ \\
\cmidrule{1-11}
\multirow[c]{4}{*}{\rot{\llamaVIII}} & \lora & E & $34.65_{\pm1.43}$ & $13.34_{\pm0.85}$ & $26.40_{\pm0.95}$ & $\underline{39.70}_{\pm0.35}$ & $70.73_{\pm0.99}$ & $38.85_{\pm0.90}$ & $4.67_{\pm2.97}$ & $16.19_{\pm0.11}$ \\
\cmidrule{2-11}
\rot{} & \multirow[t]{3}{*}{\mtllora} & E,O & $32.17_{\pm0.52}$ & $11.94_{\pm0.23}$ & $23.98_{\pm0.22}$ & $39.35_{\pm0.44}$ & $69.49_{\pm0.21}$ & $36.81_{\pm0.06}$ & $\underline{17.14}_{\pm0.98}$ & $\underline{20.01}_{\pm0.62}$ \\
\rot{} &  & E,W & $\underline{37.58}_{\pm0.12}$ & $13.68_{\pm0.05}$ & $\underline{27.02}_{\pm0.03}$ & $38.26_{\pm0.00}$ & $\underline{71.30}_{\pm0.00}$ & $\underline{38.88}_{\pm0.02}$ & $8.45_{\pm0.00}$ & $6.44_{\pm0.00}$ \\
\rot{} &  & E,O,W & $36.38_{\pm0.22}$ & $\underline{13.72}_{\pm0.07}$ & $25.75_{\pm0.23}$ & $36.19_{\pm0.00}$ & $70.94_{\pm0.04}$ & $37.24_{\pm0.17}$ & $8.76_{\pm0.13}$ & $2.04_{\pm0.05}$ \\
\cmidrule{1-11}
\multirow[c]{4}{*}{\rot{\deepseekrI}} & \lora & E & $23.89_{\pm0.27}$ & $7.57_{\pm0.30}$ & $18.48_{\pm0.30}$ & $39.34_{\pm0.32}$ & $63.60_{\pm0.24}$ & $31.49_{\pm0.34}$ & $-50.45_{\pm2.83}$ & $24.56_{\pm1.15}$ \\
\cmidrule{2-11}
\rot{} & \multirow[t]{3}{*}{\mtllora} & E,O & $26.81_{\pm1.84}$ & $8.41_{\pm0.40}$ & $19.60_{\pm1.15}$ & $39.03_{\pm0.16}$ & $65.02_{\pm0.42}$ & $32.58_{\pm1.08}$ & $\underline{-38.60}_{\pm0.76}$ & $\underline{25.44}_{\pm0.10}$ \\
\rot{} &  & E,W & $26.53_{\pm0.79}$ & $9.77_{\pm0.86}$ & $18.97_{\pm0.73}$ & $\underline{39.47}_{\pm0.49}$ & $65.37_{\pm0.23}$ & $32.14_{\pm0.83}$ & $-49.42_{\pm0.94}$ & $21.95_{\pm0.85}$ \\
\rot{} &  & E,O,W & $\underline{29.83}_{\pm0.04}$ & $\underline{11.18}_{\pm0.04}$ & $\underline{21.13}_{\pm0.07}$ & $36.58_{\pm0.00}$ & $\underline{67.35}_{\pm0.00}$ & $\underline{33.51}_{\pm0.06}$ & $-46.02_{\pm0.00}$ & $4.32_{\pm0.00}$ \\
\bottomrule
\end{tabular}

    }
}
\caption{
\textbf{Performance comparison across prompting methods (zero-shot, Chain-of-Thought, RAG) and fine-tuning strategies (LoRA, Multi-task LoRA)} on three tasks: \falcdataset (E), \orangesum (O) and \wikilargefr (W), using \mistVII, \llamaVIII and \deepseekrI models. Metrics: ROUGE-1/2/L, SARI, BERTScore-F1, composite SRB score, compression ratio, and lexical novelty. 
Results are presented as mean~$\pm$~standard deviation. Best overall results are shown in \textbf{bold}, and best results for each model are \underline{underlined}.
}
\label{table: avg results}
\end{table*}

The experimental results clearly highlight the performance benefits of both retrieval augmentation and fine-tuning approaches, particularly under multi task settings.

\paragraph{In-Context Learning (ICL).}
Zero-Shot and \COT settings generally underperform across all metrics compared to \rag\ and PEFT. While \COT\ shows a slight improvement in novelty and informativeness over Zero-Shot, gains are marginal. \rag\ consistently improves performance over basic prompting, especially on the main \falcdataset test set. For both \mistVIIinst and \llamaVIIIinst, \rag\ with task combinations (E, E+O, E+W, E+O+W) achieves substantial boosts in ROUGE and SARI scores. Notably, \rag\ yields the highest performance in most individual metrics under the ICL category.

\paragraph{Parameter-Efficient Fine-Tuning (PEFT)} models consistently outperform in-context learning (ICL) methods across all evaluation metrics. Both the \lora and \mtllora setups yield notable gains in fluency, simplicity, and informativeness. Among them, \llamaVIII-\mtllora achieves the best overall performance, excelling in metrics such as SARI, BERTScore-F1, and compression ratio, highlighting its effectiveness in producing simplified text that remains semantically faithful. The MTL-LoRA+(E+W) variant records the highest scores for SARI (44.67), BERTScore (74.05), and compression ratio (56.11), suggesting a well-balanced approach that preserves meaning while substantially reducing text length.
Additionally, we report results for the \deepseekrI model; however, its performance is consistently lower than other LLM configurations, regardless of the fine-tuning strategy applied.

\paragraph{Out-of-Domain (\falcdatasetpolitic) Performance.}
On the political subset, the performance gap between ICL and PEFT narrows slightly; however, PEFT models continue to demonstrate a clear advantage. Among the ICL methods, \rag-based approaches retain their relative lead, particularly when augmented with additional context (E+W and E+O+W), indicating stronger generalization capabilities. Notably, the zero-shot \llamaVIIIinst model achieves the highest novelty score (55.73), which could signal greater output diversity, though it might also suggest reduced fidelity. Similar to previous findings, \deepseekrI consistently underperforms compared to other LLM configurations, regardless of the fine-tuning method used.

\subsection{Human Evaluation}
\label{sec:appendix_human_evaluation}

We conduct a comprehensive human evaluation on two datasets, \falcdataset and \falcdatasetpolitic, assessing the generated explanations along dimensions guided by the ETR framework and general language quality metrics. Results are reported in Tables~\ref{tab:human-eval-results} and~\ref{tab:human-eval-quality-results}.

\paragraph{Explanation Criteria (ETR dimensions).} On \falcdataset, all methods exhibit strong performance across information selection, word selection, and sentece construction construction (scores $> 0.88$), with the \lora method slightly outperforming others in word selection\ ($0.94$) and overall global quality\ ($0.91$). Illustration quality, however, remains a consistent weakness across methods, with high variance indicating instability or inconsistent strategy for visual grounding.

For the more challenging \falcdatasetpolitic, overall scores decrease across all explanation criteria. Notably, \rag\ with joint training on E and W achieves the best global score ($0.80$), outperforming \lora and \mtllora. While \rag\ maintains high scores in information selection\ and sentece construction\, illustration scores remain low across the board, underscoring the difficulty of generating coherent examples or analogies in politically sensitive domains.

\paragraph{General Language Quality.} As shown in Table~\ref{tab:human-eval-quality-results}, \rag\ again performs competitively on both datasets. On \falcdataset, it achieves the highest ratings in grammar and coherence (both $>4.4$), with strong fluency and relevance. \mtllora slightly improves grammaticality, but this does not translate to gains in perceived overall quality.

In the political domain, quality metrics decline, consistent with the ETR scores. \rag\ trained on E and W maintains robust fluency and coherence, achieving the best overall quality score ($3.76$). In contrast, \mtllora's performance degrades notably in global quality ($2.62$), despite competitive scores in coherence and relevance, suggesting potential trade-offs introduced by multitask learning in more nuanced domains.

\paragraph{Summary.} These results highlight \rag's robustness across both explanation and linguistic quality metrics, particularly when trained jointly on ETR and sentence simplification tasks. The consistent underperformance in illustration generation across all models indicates a need for future work on grounded or multimodal explanation strategies, especially in high-stakes domains like politics.

\begin{table*}[!ht]
\tiny
\centering

\renewcommand{\arraystretch}{1}
\resizebox{\textwidth}{!}{
\begin{tabular}{LLLccccc}
\toprule
 & \textbf{Method} & \textbf{Task} & \textbf{\infochoice} & \textbf{\wordchoice} & \textbf{\sentence} & \textbf{\illustrations} & \textbf{\globalscore} \\
\midrule
\multicolumn{3}{c}{\textbf{\falcdataset}} \\ \midrule
\multirow[c]{4}{*}{\rotatebox{90}{\llamaVIII}} & \lora & E & $0.89_{\pm{0.08}}$ & $0.94_{\pm{0.04}}$ & $0.91_{\pm{0.05}}$ & $0.38_{\pm{0.40}}$ & $0.91_{\pm{0.04}}$ \\
\rotatebox{90}{} & \mtllora & E,O,W & $0.88_{\pm{0.06}}$ & $0.89_{\pm{0.07}}$ & $0.93_{\pm{0.04}}$ & $0.50_{\pm{0.65}}$ & $0.89_{\pm{0.04}}$ \\
\cmidrule{2-8}
\rotatebox{90}{} & \multirow[t]{2}{*}{\rag} & E & $0.88_{\pm{0.07}}$ & $0.92_{\pm{0.05}}$ & $0.89_{\pm{0.04}}$ & $0.40_{\pm{0.52}}$ & $0.89_{\pm{0.04}}$ \\
\rotatebox{90}{} &  & E,W & $0.91_{\pm{0.05}}$ & $0.88_{\pm{0.07}}$ & $0.92_{\pm{0.04}}$ & $0.50_{\pm{0.44}}$ & $0.89_{\pm{0.04}}$ \\
\midrule
\multicolumn{3}{c}{\textbf{\falcdatasetpolitic}} \\ \midrule
\multirow[c]{4}{*}{\rotatebox{90}{\llamaVIII}} & \lora & E & $0.77_{\pm{0.14}}$ & $0.66_{\pm{0.11}}$ & $0.79_{\pm{0.11}}$ & $0.15_{\pm{0.24}}$ & $0.73_{\pm{0.08}}$ \\
\rotatebox{90}{} & \mtllora & E,O,W & $0.69_{\pm{0.13}}$ & $0.59_{\pm{0.11}}$ & $0.65_{\pm{0.12}}$ & $0.27_{\pm{0.27}}$ & $0.64_{\pm{0.08}}$ \\
\cmidrule{2-8}
\rotatebox{90}{} & \multirow[t]{2}{*}{\rag} & E & $0.82_{\pm{0.09}}$ & $0.74_{\pm{0.10}}$ & $0.86_{\pm{0.07}}$ & $0.10_{\pm{0.23}}$ & $0.78_{\pm{0.05}}$ \\
\rotatebox{90}{} &  & E,W & $0.87_{\pm{0.06}}$ & $0.75_{\pm{0.09}}$ & $0.85_{\pm{0.08}}$ & $0.40_{\pm{0.37}}$ & $0.80_{\pm{0.06}}$ \\
\bottomrule
\end{tabular}
}
\caption{
\textbf{Human evaluation of generations based on ETR guideline criteria}, comparing various methods on the \falcdataset and \falcdatasetpolitic test sets using their optimal ICL and MTL configurations. 
Each method is evaluated along four explanation dimensions: {\infochoice} (information selection), {\wordchoice} (lexical choice), {\sentence} (sentence construction), \illustrations, and \globalscore\ representing the overall quality score. 
Training tasks are abbreviated as E (\falcdataset), O (\orangesum), and W (\wikilargefr). 
Reported scores are means with 95\% confidence intervals.
}
\label{tab:human-eval-results}
\end{table*}

\begin{table*}[!ht]

\renewcommand{\arraystretch}{1.1}
\tiny
\centering
\resizebox{\textwidth}{!}{\begin{tabular}{LLLccccc}
\toprule
 & \textbf{Method} & \textbf{Task} & \textbf{\fluency} & \textbf{\grammar} & \textbf{\relevance} & \textbf{\coherence} & \textbf{\quality} \\
\midrule
\multicolumn{3}{c}{\textbf{\falcdataset}} \\ \midrule
\multirow[c]{4}{*}{\rotatebox{90}{\llamaVIII}} & \lora & E & $4.29_{\pm{0.26}}$ & $4.57_{\pm{0.23}}$ & $3.95_{\pm{0.39}}$ & $4.24_{\pm{0.32}}$ & $3.95_{\pm{0.37}}$ \\
\rotatebox{90}{} & \mtllora & E,O,W & $4.33_{\pm{0.33}}$ & $4.67_{\pm{0.22}}$ & $4.10_{\pm{0.38}}$ & $4.14_{\pm{0.39}}$ & $3.95_{\pm{0.44}}$ \\
\cmidrule{2-8}
\rotatebox{90}{} & \multirow[t]{2}{*}{\rag} & E & $4.43_{\pm{0.27}}$ & $4.71_{\pm{0.21}}$ & $4.24_{\pm{0.38}}$ & $4.43_{\pm{0.34}}$ & $4.24_{\pm{0.35}}$ \\
\rotatebox{90}{} &  & E,W & $4.43_{\pm{0.23}}$ & $4.57_{\pm{0.23}}$ & $4.43_{\pm{0.34}}$ & $4.52_{\pm{0.27}}$ & $3.95_{\pm{0.34}}$ \\
\midrule
\multicolumn{3}{c}{\textbf{\falcdatasetpolitic}} \\ \midrule
\multirow[c]{4}{*}{\rotatebox{90}{\llamaVIII}} & \lora & E & $3.90_{\pm{0.52}}$ & $4.43_{\pm{0.42}}$ & $4.24_{\pm{0.43}}$ & $4.24_{\pm{0.45}}$ & $3.14_{\pm{0.62}}$ \\
\rotatebox{90}{} & \mtllora & E,O,W & $3.81_{\pm{0.45}}$ & $4.48_{\pm{0.34}}$ & $4.40_{\pm{0.38}}$ & $4.52_{\pm{0.23}}$ & $2.62_{\pm{0.55}}$ \\
\cmidrule{2-8}
\rotatebox{90}{} & \multirow[t]{2}{*}{\rag} & E & $4.24_{\pm{0.38}}$ & $4.48_{\pm{0.34}}$ & $4.10_{\pm{0.35}}$ & $4.33_{\pm{0.30}}$ & $3.45_{\pm{0.44}}$ \\
\rotatebox{90}{} &  & E,W & $4.33_{\pm{0.33}}$ & $4.57_{\pm{0.23}}$ & $4.29_{\pm{0.29}}$ & $4.43_{\pm{0.27}}$ & $3.76_{\pm{0.40}}$ \\
\bottomrule
\end{tabular}}
\caption{
\textbf{Human ratings of fluency, grammar, relevance, coherence, and overall quality} for different methods evaluated on the \falcdataset and \falcdatasetpolitic test sets, using their optimal ICL and MTL configurations. 
Training tasks are abbreviated as E (\falcdataset), O (\orangesum), and W (\wikilargefr). 
Scores are reported as means with 95\% confidence intervals.
}
\label{tab:human-eval-quality-results}
\end{table*}

\subsection{Comparison of Ground Truth and Generated ETR Outputs}

The table~\ref{tab:generation vs gt} presents a detailed comparison of different model configurations (Mistral-7B and LlaMA-8B), training methods (RAG, LoRA, MTL-LoRA), and task combinations (ETR, summarization and simplification). Metrics include the average number of words and sentences, sentence length, KMRE (higher is better), novelty, and compression ratio.

Overall, models trained with MTL-LoRA tend to generate more concise outputs while maintaining strong performance in terms of KMRE. For instance, LlaMA-8B + MTL-LoRA (E,W) achieves the highest KMRE score (102.98) and the highest novelty (33.05), indicating its ability to produce informative and diverse content.

RAG-based methods generally generate longer texts, with higher sentence lengths (up to 11.07 words on average for {LlaMA-8B + RAG (E,O,W)}), but often at the expense of novelty. This suggests that RAG relies more heavily on retrieved content, which may reduce the originality of generated text.

Compared to the ground truth, the generated texts generally contain more words and exhibit equal or greater sentence lengths. Notably, the MTL-LoRA configurations achieve higher compression ratios, highlighting their ability to effectively condense information. While no method fully replicates the characteristics of the test set, defined by its notably short sentences and high compression. LlaMA-8B MTL-LoRA trained on Wikilarge (W) and ETR-fr (E) yields outputs that most closely resemble the test set in terms of both compression and sentence structure.

\begin{table*}[!ht]
\centering
\small
\renewcommand{\arraystretch}{1.1}
\begin{tabular}{LLLccccccc}
\toprule
 & Method & Tasks & \textbf{\# Words} & \textbf{\# Sentences} & \textbf{Sentence length} & \textbf{KMRE $\uparrow$} & \textbf{Novelty} & \textbf{Comp. ratio} \\

\midrule

\multicolumn{2}{c}{\multirow[c]{1}{*}{\textbf{\textit{Ground Truth}}}} & \multicolumn{1}{l}{\textbf{\textit{Test Set}}} & 40.26 & 8.91 & 4.64 & 102.99 & 55.01 & 65.19 \\

\midrule
\multirow[c]{4}{*}{\rotatebox{90}{Mistral-7B}} & \multirow[t]{4}{*}{RAG} & E & 66.38 & 7.70 & 8.76 & 99.77 & 26.55 & 44.32 \\
\rot{} &  & E,O & 60.91 & 6.13 & 10.05 & 97.21 & 26.61 & 48.45 \\
\rot{} &  & E,W & 80.74 & 7.83 & 10.67 & 97.37 & 23.01 & 33.80 \\
\rot{} &  & E,O,W & 62.45 & 6.15 & 10.25 & 97.62 & 25.85 & 46.42 \\
\midrule

\multirow[c]{4}{*}{\rot{LlaMA-8B}} & \multirow[t]{4}{*}{RAG} & E & 63.72 & 7.87 & 8.38 & 101.70 & 27.14 & 46.18 \\
\rot{} &  & E,O & 74.19 & 7.57 & 9.92 & 97.45 & 24.29 & 39.22 \\
\rot{} &  & E,W & 69.72 & 7.64 & 9.49 & 100.34 & 25.26 & 41.89 \\
\rot{} &  & E,O,W & 87.17 & 8.40 & 11.07 & 97.48 & 23.69 & 25.94 \\
\midrule
\multirow[c]{4}{*}{\rot{Mistral-7B}} & LoRA & E & 65.55 & 9.26 & 7.73 & 101.20 & 18.35 & 44.42 \\
\cmidrule{2-9}
\rot{} & \multirow[t]{3}{*}{MTL-LoRA} & E,O & 56.75 & 8.25 & 7.38 & 102.61 & 24.17 & 53.48 \\
\rot{} &  & E,W & 54.08 & 9.28 & 6.46 & 104.23 & 24.99 & 53.62 \\
\rot{} &  & E,O,W & 60.08 & 8.81 & 7.23 & 101.80 & 23.38 & 48.93 \\
\midrule
\multirow[c]{4}{*}{\rot{LlaMA-8B}} & LoRA & E & 56.96 & 8.64 & 7.62 & 100.93 & 18.87 & 50.66 \\
\cmidrule{2-9}
\rot{} & \multirow[t]{3}{*}{MTL-LoRA} & E,O & 60.08 & 9.87 & 7.00 & 100.84 & 23.06 & 51.36 \\
\rot{} &  & E,W & 50.09 & 9.19 & 6.50 & 102.98 & 33.05 & 56.11 \\
\rot{} &  & E,O,W & 54.06 & 8.77 & 7.42 & 101.35 & 24.39 & 53.24 \\
\bottomrule
\end{tabular}
\caption{Comparison of different model configurations (Mistral-7B and LlaMA-8B) and training methods (RAG, LoRA, MTL-LoRA) across various task combinations (E: \falcdataset, O: \orangesum, W: \wikilargefr). The metrics include word count, sentence count, average sentence length, KMRE (higher is better), novelty, and compression ratio. Ground truth statistics from the test set are also provided for reference.}
\label{tab:generation vs gt}
\end{table*}
\section{In-Context Learning Hyperparameters Effects}
\label{section:prompts}
\begin{figure}[!ht]
    \centering
    \begin{subfigure}{\linewidth}
        \centering
        \includegraphics[width=\linewidth]{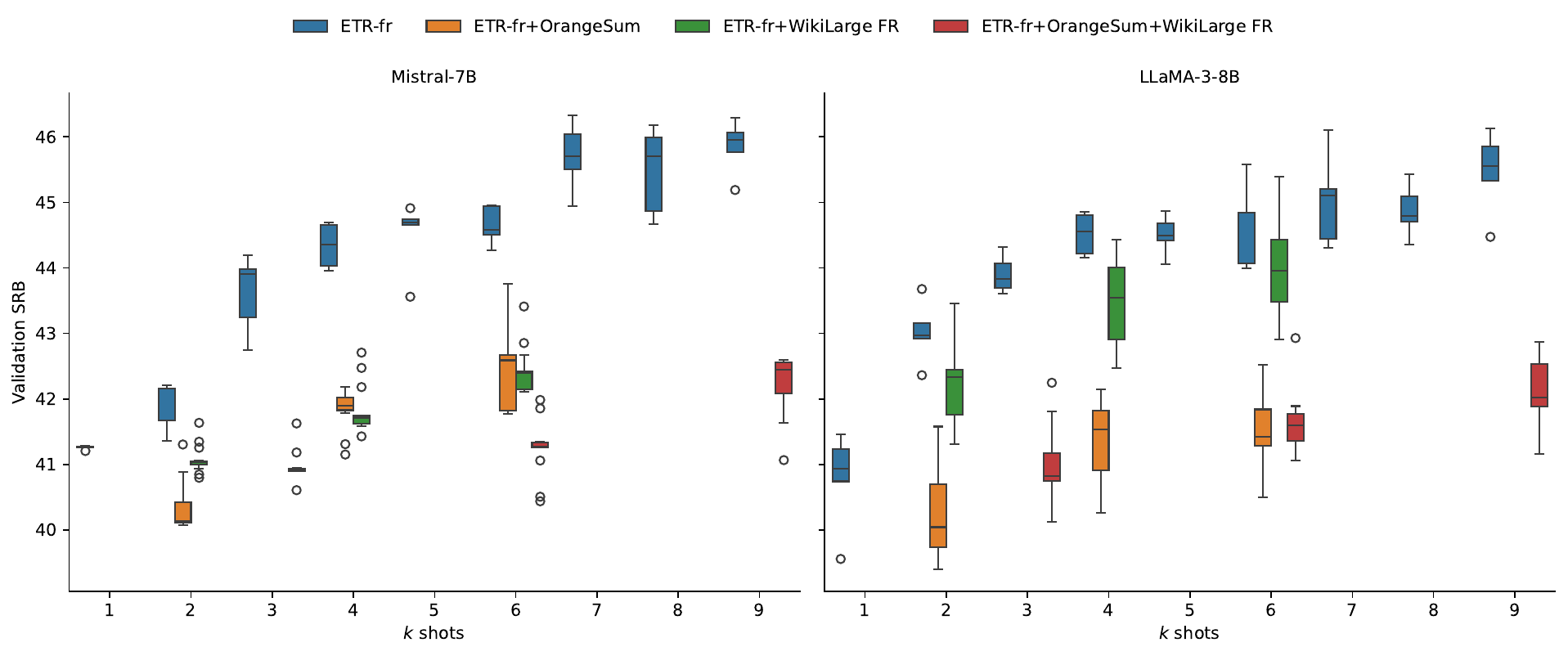}
        \caption{SRB performance under varying number of in-context examples ($k \in [1;9]$) and task combinations.}
        \label{fig:k_shots}
    \end{subfigure}%
    \hfill
    \begin{subfigure}{\linewidth}
        \centering
        \includegraphics[width=\linewidth]{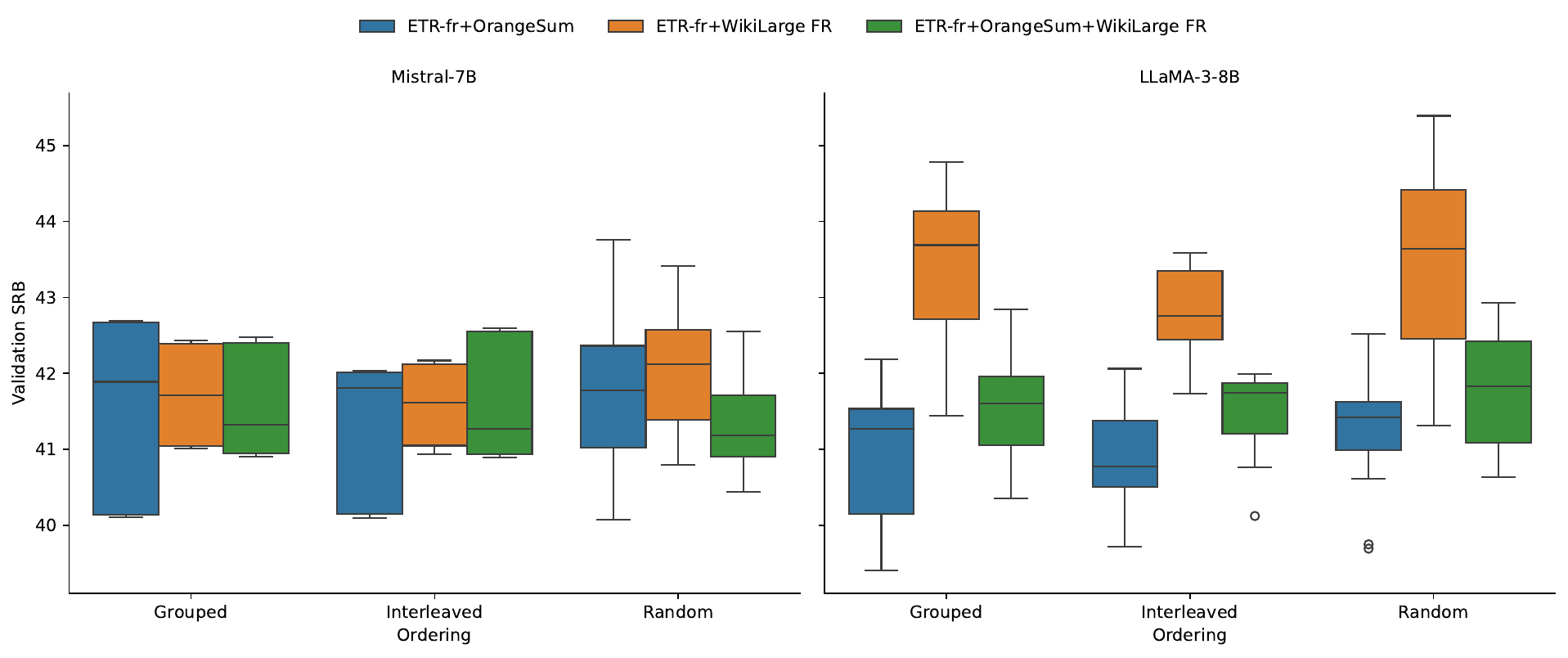}
        \caption{SRB performance under different example ordering strategies and task combination configurations.}
        \label{fig:task_ordering}
    \end{subfigure}
    \caption{Comparison of SRB performance on the ETR-fr validation set across different in-context settings and ordering strategies.}
    \label{fig:rag_analysis}
\end{figure}

Figure~\ref{fig:prompt_comparison} illustrates examples of prompts used for zero-shot (Fig.~\ref{fig:prompt_zero_shot}), chain-of-thought (Fig.~\ref{fig:prompt_cot}) and few-shot (Fig.~\ref{fig:prompt_few_shot}).

\subsection{Impact of the Number of Shots on ETR-fr Performance}
Figure~\ref{fig:k_shots} presents the performance of LLaMA-3-8B and Mistral-7B on the French text simplification benchmark (ETR-fr) across varying numbers of in-context learning (ICL) examples ($k = 1$ to $9$) and under different training configurations.

\paragraph{LLaMA-3-8B Performance.}
For the {LLaMA-3-8B} model, performance generally increases with larger $k$ values. The basic task {ETR-fr} alone yields steadily rising median SRB scores, from 40.93 at $k=1$ to 45.96 at $k=9$. The incorporation of auxiliary datasets ({OrangeSum} and {WikiLarge FR}) leads to varied results. For instance, combining {ETR-fr} with {WikiLarge FR} at $k=2$ raises the median from 42.96 to 42.33, while the three-dataset combination at $k=6$ has a lower median of 41.60 compared to 44.84 for {ETR-fr} alone. This suggests diminishing returns or even negative interference when too many tasks are combined.

\paragraph{Mistral-7B Performance.}
The {Mistral-7B} model demonstrates a similar trend of improved performance with increasing $k$ values for the {ETR-fr} task. Median SRB rise from 41.26 at $k=1$ to 45.96 at $k=9$. However, Mistral exhibits less variation across configurations. The inclusion of {OrangeSum} and {WikiLarge FR} improves SRB modestly, and the three-dataset combination remains slightly below the single-task performance. For example, at $k=6$, {ETR-fr} alone achieves a median of 44.58, whereas the triple combination achieves only 41.28.

\paragraph{Comparative Insights.}
When comparing both models, {LLaMA-3-8B} tends to show greater gains from dataset combinations than {Mistral-7B}, although it also experiences more variance. For both models, the highest performances are obtained when using {ETR-fr} alone at higher $k$ values, indicating that overloading the prompt context with multiple tasks may dilute performance. Moreover, the higher maximum SRB for LLaMA across configurations (e.g., up to 46.12) suggest it may have a higher performance ceiling, but with more fluctuation.


\subsection{Impact of the Tasks Ordering on ETR-fr Performance}

Figure~\ref{fig:task_ordering} presents the impact of task ordering on model performance under different multi-task training configurations.
For both models, three types of example ordering are compared: \textit{grouped}, \textit{interleaved}, and \textit{random}. Each ordering is evaluated with different training task combinations, such as {ETR-fr+OrangeSum}, {ETR-fr+WikiLarge FR}, and {ETR-fr+OrangeSum+WikiLarge FR}.

\paragraph{LLaMA-3-8B Performance.}
For {LLaMA-3-8B}, performance consistently improves when WikiLarge FR data is added to the training set. The configuration using only {ETR-fr+WikiLarge FR} yields the highest SRB scores across all ordering methods, particularly under the {random} strategy, which achieves the highest maximum score (45.39). Overall, {grouped} and {random} orderings tend to result in higher median and upper-quartile SRB compared to {interleaved} ordering, indicating that the sequential arrangement of examples plays a role in performance.

\paragraph{Mistral-7B Performance.}
For {Mistral-7B}, the impact of training set composition is similarly positive, with improvements observed upon including WikiLarge FR. However, the differences among the three ordering strategies are more subtle. {grouped} and {interleaved} yield very similar statistics, with slight advantages in median SRB depending on the training data. The highest maximum score for Mistral-7B (43.76) occurs under the {random} strategy with the {ETR-fr+OrangeSum} dataset, although this configuration does not have the most consistent results across runs.

\paragraph{Comparative Insights.}
Comparing the two models, LLaMA-3-8B generally outperforms Mistral-7B in terms of median and maximum SRB, particularly when trained with ETR-fr and WikiLarge FR. Mistral-7B demonstrates more stable performance with narrower score ranges but slightly lower central tendency metrics. These results suggest that while both models benefit from enriched prompts, LLaMA-3-8B exhibits greater potential for high-end performance when paired with appropriate example ordering and task combinations.

\definecolor{darkblue}{RGB}{46,25, 110}

\newcommand{\dssectionheader}[1]{%
   \noindent\framebox[\columnwidth]{%
      {\fontfamily{phv}\selectfont \textbf{\textcolor{darkblue}{#1}}}
   }
}

\newcommand{\dsquestion}[1]{%
    {{\footnotesize\noindent \fontfamily{phv}\selectfont \textcolor{darkblue}{\textbf{#1}}}}
}

\newcommand{\dsquestionex}[2]{%
    {{\footnotesize\noindent \fontfamily{phv}\selectfont \textcolor{darkblue}{\textbf{#1} #2}}}
}

\newcommand{\dsanswer}[1]{%
   {\noindent #1 \medskip}
}

\section{\falcdataset Dataset Sheet}
\label{apdx: etr-fr dataset sheet}


The dataset description follows the recommendations and template proposed by \citet{gebru_datasetsheet_2021}.

\bigskip
\dssectionheader{Motivation}

\dsquestionex{For what purpose was the dataset created?}{}

\dsanswer{
The ETR-fr dataset was created to address the lack of high-quality, document-aligned corpora suitable for generating Easy-to-Read (ETR) text. It supports the task of generating cognitively accessible texts for individuals with cognitive impairments by providing paragraph-aligned text pairs that follow the European ETR guidelines. This dataset enables the training and evaluation of automatic systems for ETR generation in French, targeting the linguistic and cognitive accessibility requirements typically overlooked by existing simplification or summarization.
}

\dsquestion{Who created this dataset (e.g., which team, research group) and on behalf of which entity (e.g., company, institution, organization)?}

\dsanswer{
The dataset was constructed by the authors of the this paper on ETR-fr. 
}



\bigskip
\dssectionheader{Composition}

\dsquestionex{What do the instances that comprise the dataset represent (e.g., documents, photos, people, countries)?}{ 
}

\dsanswer{
Each instance in the ETR-fr dataset consists of a pair of paragraph-aligned French texts: a source text and its corresponding Easy-to-Read (ETR) version. These are designed to support document-level simplification, emphasizing both lexical and structural transformation.
}

\dsquestion{How many instances are there in total (of each type, if appropriate)?}{}

\dsanswer{
The dataset contains 523 paragraph-aligned text pairs. Additionally, an out-of-domain subset, ETR-fr-politic, includes 33 paragraph pairs from 2022 French presidential election programs.
}

\dsquestionex{What data does each instance consist of? “Raw” data (e.g., unprocessed text or images) or features?}
{
}
\begin{figure}[!h]
    \centering
    \begin{minipage}{0.5\columnwidth}
        \fbox{\includegraphics[width=.90\textwidth,page=1]{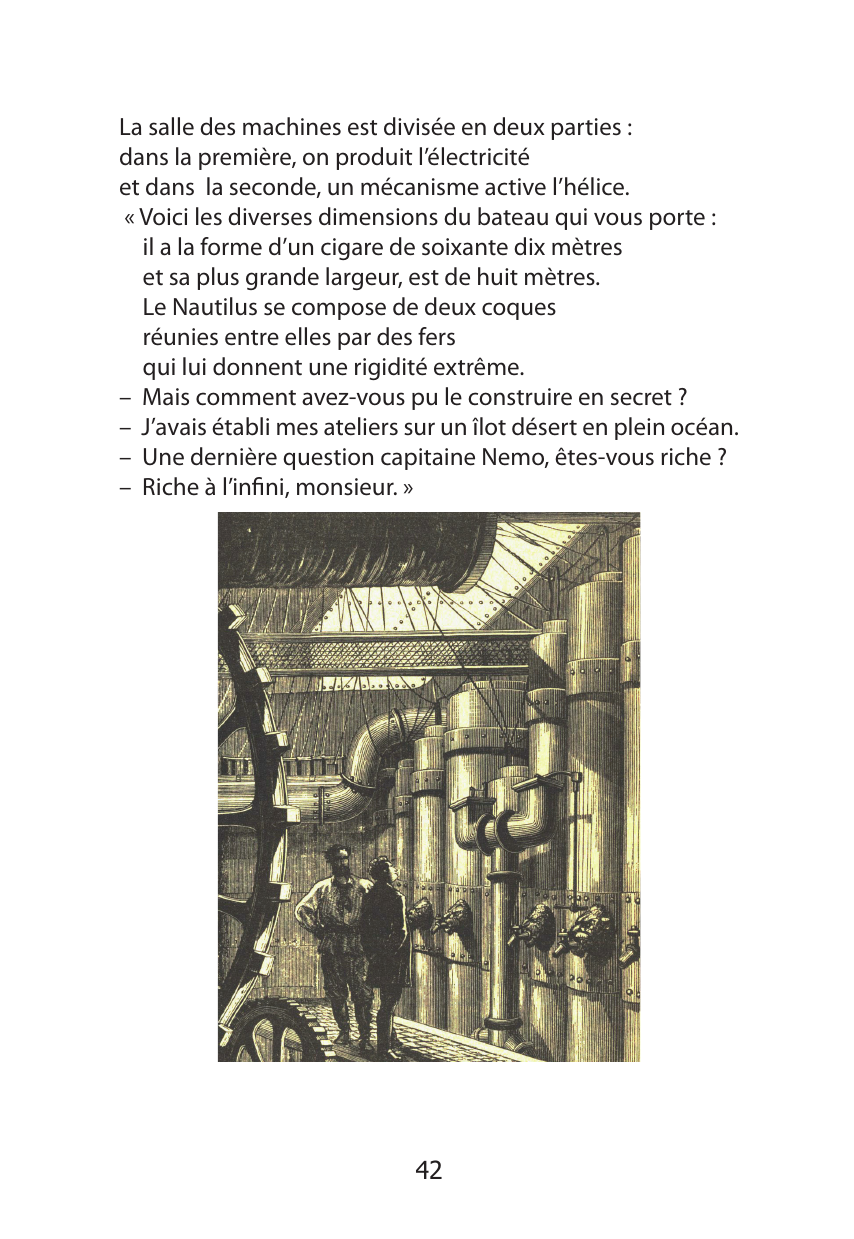}}
    \end{minipage}%
    \begin{minipage}{0.5\columnwidth}
        \fbox{\includegraphics[width=.90\linewidth,page=2]{imgs/nautilus_fr.pdf}}
    \end{minipage}
    
    \caption{Extract of the ETR book \textit{Twenty Thousand Leagues Under the Seas} by Jules Verne from François~Baudez Publishing.
    \textbf{Left page} is the original text with an illustration. 
    \textbf{Right page} is the ETR transcription with the main information plus its captioned {\it vignettes}.
    }
    \label{fig:example_etr}
\end{figure}

\dsanswer{
Each instance consists of “raw” French text paragraphs: a complex source text and its corresponding simplified (ETR) version. These are aligned at the paragraph level and include natural language text only.
}

\dsquestionex{Is there a label or target associated with each instance?}{
}

\dsanswer{
Yes. The target is the simplified (ETR-compliant) version of the source paragraph, forming a supervised text-to-text pair for generation tasks.
}

\dsquestionex{Is any information missing from individual instances?}{
}

\dsanswer{
The pictograms present with the original texts have not been extracted.
}

\dsquestionex{Are relationships between individual instances made explicit (e.g., users’ movie ratings, social network links)?}
{
}

\dsanswer{
No such relationships exist or are made explicit in this dataset.
}

\dsquestionex{Are there recommended data splits (e.g., training, development/validation, testing)?}
{
}

\dsanswer{
Yes. The dataset is divided into training (399 pairs), validation (71 pairs), and test (53 pairs) subsets. The test set comprises two distinct books chosen to ensure diversity in linguistic features such as text length, structure, and readability. The remaining books were split into training and validation sets using a stratified approach to minimize thematic and lexical overlap. Additionally, the ETR-fr-politic test set (33 pairs) was introduced to assess model generalization on out-of-domain content not seen during training.
}

\dsquestionex{Are there any errors, sources of noise, or redundancies in the dataset?}
{
}

\dsanswer{
No specific mention of noise or redundancy issues is made in the source document.
}

\dsquestionex{
Is the dataset self-contained, or does it link to or otherwise rely on external resources (e.g., websites, tweets, other datasets)?}
{
}

\dsanswer{
The dataset is self-contained, it does not rely on external resources.
}

\dsquestionex{Does the dataset contain data that might be considered confidential (e.g., data that is protected by legal privilege or by doctor-patient confidentiality, data that includes the content of individuals non-public communications)?}
{
}

\dsanswer{
No. All texts are from published sources and are intended for public consumption.
}

\dsquestionex{Does the dataset contain data that, if viewed directly, might be offensive, insulting, threatening, or might otherwise cause anxiety?}
{
}

\dsanswer{
No such content is reported or expected in the dataset.
}

\dsquestionex{Does the dataset relate to people?}
{
}

\dsanswer{
No. The dataset is composed of literary and political texts and does not contain personal information.
}

\bigskip
\dssectionheader{Collection Process}

\dsquestionex{How was the data associated with each instance acquired?}{
}

\dsanswer{
The data are directly observable from published ETR books. Each ETR version is produced by a pair of trained transcribers working collaboratively, in accordance with the European Easy-to-Read guidelines~\cite{pathways_information_2021}, to obtain official ETR certification.
}

\dsquestionex{What mechanisms or procedures were used to collect the data (e.g., hardware apparatus or sensor, manual human curation, software program, software API)?}
{
}

\dsanswer{
To collect the data from ETR books, we first obtained the PDF versions and manually curated them to identify pairs of pages containing the original text and its corresponding ETR version. The textual content was then extracted using the Python library \texttt{pypdfium2}\footnote{\url{https://github.com/pypdfium2-team/pypdfium2}}.
}

\dsquestion{If the dataset is a sample from a larger set, what was the sampling strategy (e.g., deterministic, probabilistic with specific sampling probabilities)?}

\dsanswer{
The dataset is not sampled from a larger set; it includes the complete collection of available aligned texts selected for the study.
}

\dsquestion{Who was involved in the data collection process (e.g., students, crowdworkers, contractors) and how were they compensated (e.g., how much were crowdworkers paid)?}

\dsanswer{
Unknown for the mannual book transcrptions.
The data collection was carried out by the main author of this paper as part of their research work.
}

\dsquestionex{Over what timeframe was the data collected? Does this timeframe match the creation timeframe of the data associated with the instances (e.g., recent crawl of old news articles)?}
{
}

\dsanswer{
The exact creation dates of the original books are unknown. However, the dataset itself was constructed between May 2023 and June 2023.
}

\dsquestionex{Were any ethical review processes conducted (e.g., by an institutional review board)?}
{
}

\dsanswer{
No ethical review.
}

\dsquestionex{Does the dataset relate to people?}
{
}

\dsanswer{
No.
}

\bigskip
\dssectionheader{Preprocessing/cleaning/labeling}

\dsquestionex{Was any preprocessing/cleaning/labeling of the data done (e.g., discretization or bucketing, tokenization, part-of-speech tagging, SIFT feature extraction, removal of instances, processing of missing values)?}
{
}

\dsanswer{
Manual cleaning was performed to remove chapter titles from the original texts, as these were not present in the corresponding ETR versions.
}

\dsquestionex{Was the “raw” data saved in addition to the preprocessed/cleaned/labeled data (e.g., to support unanticipated future uses)?}
{
}

\dsanswer{
Yes. The raw data is provided alongside the cleaned version.
}

\dsquestionex{Is the software used to preprocess/clean/label the instances available?}
{
}

\dsanswer{
\begin{itemize}
    \item \texttt{pypdfium2}: \url{https://github.com/pypdfium2-team/pypdfium2}
    \item \texttt{cleantext}: \url{https://pypi.org/project/cleantext/}
\end{itemize}
}

\bigskip
\dssectionheader{Uses}

\dsquestionex{Has the dataset been used for any tasks already?}{
}

\dsanswer{
No.
}

\dsquestion{What (other) tasks could the dataset be used for?}

\dsanswer{
This dataset could also be used for text classification and style transfer.
}

\dsquestionex{Is there anything about the composition of the dataset or the way it was collected and preprocessed/cleaned/labeled that might impact future uses?}
{
}

\dsanswer{
No. 
}

\dsquestionex{Are there tasks for which the dataset should not be used?}
{
}

\dsanswer{
No.
}

\bigskip
\dssectionheader{Distribution}


%

\dsquestionex{How will the dataset will be distributed (e.g., tarball on website, API, GitHub)}
{
}

\dsanswer{
The dataset will be available on GitHub repository.
}

\dsquestion{When will the dataset be distributed?}

\dsanswer{
The dataset will be released pending agreement from the ETR books publisher.
}

\dsquestionex{Will the dataset be distributed under a copyright or other intellectual property (IP) license, and/or under applicable terms of use (ToU)?}
{
}

\dsanswer{
The dataset will be released under a custom license, subject to approval from the ETR books publisher. Redistribution and use will be permitted for research purposes only, with appropriate citation. No commercial use will be allowed without explicit permission.
}

\dsquestionex{Have any third parties imposed IP-based or other restrictions on the data associated with the instances?}
{
}

\dsanswer{
No.
}

\dsquestionex{Do any export controls or other regulatory restrictions apply to the dataset or to individual instances?}
{
}

\dsanswer{
No restrictions.
}

\bigskip
\dssectionheader{Maintenance}

\dsquestion{Who will be supporting/hosting/maintaining the dataset?}

\dsanswer{
The dataset will be maintained by the primary author of the paper.
}

\dsquestion{How can the owner/curator/manager of the dataset be contacted (e.g., email address)?}

\dsanswer{
By submitting an issue on the dataset's GitHub repository.
}

\dsquestionex{Is there an erratum?}
{
}

\dsanswer{
Yes, errata can be reported and tracked via GitHub issues.
}

\dsquestionex{Will the dataset be updated (e.g., to correct labeling errors, add new instances, delete instances)?}
{
}

\dsanswer{
Yes, updates will be handled by the repository maintainer on GitHub. Users can receive update notifications by subscribing to the repository.
}

\dsquestionex{If the dataset relates to people, are there applicable limits on the retention of the data associated with the instances (e.g., were individuals in question told that their data would be retained for a fixed period of time and then deleted)?}
{
}

\dsanswer{
This dataset does not contain or pertain to any personal data.
}

\dsquestionex{Will older versions of the dataset continue to be supported/hosted/maintained?}
{
}

\dsanswer{
Yes, previous versions will remain available in the “Releases” section of the GitHub repository.
}

\dsquestionex{If others want to extend/augment/build on/contribute to the dataset, is there a mechanism for them to do so?}
{
}

\dsanswer{
Yes, contributors may open a GitHub issue and submit a pull request. They should mention the maintainer and clearly describe their proposed changes, which will then be reviewed and validated before being merged.
}
\section{Human Evaluation Questions}

Table~\ref{tab:questions} presents a comprehensive set of human evaluation questions based on the ETR European guidelines, organized into four key categories: Information Choice, Sentence Construction and Word Choice, Illustrations, and Overall Quality. Each category includes multiple criteria designed to assess the clarity, structure, and accessibility of information provided in a text. For example, the Information Choice section evaluates whether essential information is prioritized, logically ordered, and clearly grouped. Sentence Construction and Word Choice emphasizes linguistic simplicity, clarity, and consistency, discouraging complex vocabulary, metaphors, or abbreviations unless adequately explained. The Illustrations section assesses the use of relatable examples to clarify abstract ideas, while the Quality section covers fluency, grammar, factual correctness, coherence, and other aspects of textual integrity. These criteria serve as a structured framework to ensure texts are understandable, reader-friendly, and fit for purpose.

\begin{table*}[!ht]
    \centering
    \small
\begin{tabularx}{\textwidth}{|l|l|X|}
\hline
\textbf{Information Choice} & \textbf{Code} & \textbf{Description} \\
\hline
\multirow{5}{3.5cm}{Information Choice} 
& CI3 & Providing too much information can create confusion. Only important information should be given. Is this criterion met? \\
\cline{2-3}
 & CI4 & Are the pieces of information placed in an order that is easy to follow and understand? \\
\cline{2-3}
 & CI5 & Is the main information easy to find? \\
\cline{2-3}
 & CI6 & Are pieces of information about the same topic grouped together? \\
\cline{2-3}
 & CI8 & Are important pieces of information repeated? \\
\hline
\multirow{21}{3.5cm}{Sentence construction and word choice} 
& CPM1 & Are the sentences short? \\
\cline{2-3}
 & CPM2 & Are the words easy to understand? \\
\cline{2-3}
 & CPM3 & Are difficult words clearly explained when you use them? \\
\cline{2-3}
 & CPM4 & Are difficult words explained more than once? \\
\cline{2-3}
 & CPM5 & Is the language used the most suitable for the people who will use the information? \\
\cline{2-3}
 & CPM6 & Is the same word used throughout the document to describe the same thing? \\
\cline{2-3}
 & CPM7 & Difficult and abstract ideas like metaphors should not be used. Is this criterion met? \\
\cline{2-3}
 & CPM8 & Uncommon words in a foreign language should not be used. Is this criterion met? \\
\cline{2-3}
 & CPM9 & Contracted words, like text messaging slang, should not be used. Is this criterion met? \\
\cline{2-3}
 & CPM10 & Does the author address directly the people for whom the information is intended? \\
\cline{2-3}
 & CPM11 & Can you easily identify to whom or what the pronouns correspond? \\
\cline{2-3}
 & CPM12 & Are positive sentences rather than negative ones used whenever possible? \\
\cline{2-3}
 & CPM13 & Is the active voice used instead of the passive voice whenever possible? \\
\cline{2-3}
 & CPM14 & Is the punctuation simple? \\
\cline{2-3}
 & CPM15 & Are bullets or numbers used instead of lists of words separated by commas? \\
\cline{2-3}
 & CPM16 & Are numbers written in digits (1, 2, 3) rather than words? \\
\cline{2-3}
 & CPM17 & Acronyms should be avoided or explained when used. Is this criterion met? \\
\cline{2-3}
 & CPM18 & Abbreviations should not be used. Is this criterion met? \\
\cline{2-3}
 & CPM19 & Are dates written out in full? \\
\cline{2-3}
 & CPM20 & The use of percentages or large numbers should be limited and always explained. Is this criterion met? \\
\cline{2-3}
 & CPM21 & Special characters should not be used. Is this criterion met? \\
\hline
\multirow{2}{3.5cm}{Illustrations} 
& I1 & Are there examples to illustrate complex ideas? \\
\cline{2-3}
 & I2 & Are examples, as much as possible, drawn from everyday life? \\
\hline
\multirow{8}{3.5cm}{Quality} 
& CA1 & Language fluency \\
\cline{2-3}
 & CA2 & Grammar / Spelling \\
\cline{2-3}
 & CA3 & Factual accuracy \\
\cline{2-3}
 & CA4 & Textual coherence \\
\cline{2-3}
 & CA5 & Presence of copies from the original text? \\
\cline{2-3}
 & CA6 & Presence of chaotic repetitions? \\
\cline{2-3}
 & CA7 & Presence of hallucinations? \\
\cline{2-3}
 & CA8 & Overall perceived quality \\
\hline
\end{tabularx}
\caption{Evaluation criteria, extracted from ETR European guidelines, for information clarity, sentence construction, illustrations, and quality.}
\label{tab:questions}
\end{table*}
\begin{figure*}[!ht]
    \centering
    \scriptsize
    \begin{subfigure}[b]{1.0\textwidth}
        \centering
        \input{latex/prompts/prompt.zero_shot}
        \caption{Zero Shot Prompt}
        \label{fig:prompt_zero_shot}
    \end{subfigure}
    \hfill
    \begin{subfigure}[b]{1.0\textwidth}
        \centering
        \input{latex/prompts/prompt.few_shot}
        \caption{Few Shot Prompt}
        \label{fig:prompt_few_shot}
    \end{subfigure}
\end{figure*}

\begin{figure*}\ContinuedFloat
    \scriptsize
    \begin{subfigure}[b]{1.0\textwidth}
        \centering
        \input{latex/prompts/prompt.cot}
        \caption{Chain of Thought Prompt}
        \label{fig:prompt_cot}
    \end{subfigure}

    \caption{Zero Shot, Chain of Thought and Few Shot Prompts}
    \label{fig:prompt_comparison}
\end{figure*}

\end{document}